\documentclass[pdflatex,sn-mathphys]{sn-jnl}

\usepackage{graphicx}
\usepackage{dsfont}
\usepackage{amsmath}
\usepackage{amsfonts}
\usepackage{subcaption}
\usepackage{booktabs}
\usepackage{float}

\jyear{2022}%

\begin{document}

\pdfpagewidth16cm
\pdfpageheight24cm

\title[Quantifying Quality of Class-Conditional Generative Models]{Quantifying Quality of Class-Conditional Generative Models in Time-Series Domain}


\author*[1,2,3]{\fnm{Alireza} \sur{Koochali}}\email{alireza.koochali@iav.de}
\author[1,3]{\fnm{Maria} \sur{Walch}}\email{maria.walch@iav.de}
\author[]{\fnm{Sankrutyayan} \sur{Thota}}\email{sankrutyayan@gmail.com}
\author[1]{\fnm{Peter} \sur{Schichtel}}\email{peter.schichtel@iav.de}
\author[2,3]{\fnm{Andreas} \sur{Dengel}}\email{andreas.dengel@dfki.de}
\author[2]{\fnm{Sheraz} \sur{Ahmed}}\email{sheraz.ahmed@dfki.de}

\affil[1]{\orgname{IAV GmbH}, \orgaddress{\street{Trippstadterstr. 122}, \city{Kaiserslautern}, \postcode{67663}, \country{Germany}}}

\affil[2]{\orgname{DFKI GmbH}, \orgaddress{\street{Trippstadterstr. 122}, \city{Kaiserslautern}, \postcode{67663}, \country{Germany}}}

\affil[3]{\orgdiv{Computer Science Department}, \orgname{University of Kaiserslautern}, \orgaddress{\street{Erwin-Schr\"oingerstr 52}, \city{Kaiserslautern}, \postcode{67663}, \country{Germany}}}


\abstract{Generative models are designed to address the data scarcity problem. Even with the exploding amount of data, due to computational advancements, some applications (e.g., health care, weather forecast, fault detection) still suffer from data insufficiency, especially in the time-series domain. Thus generative models are essential and powerful tools, but they still lack a consensual approach for quality assessment. Such deficiency hinders the confident application of modern implicit generative models on time-series data.
Inspired by assessment methods on the image domain, we introduce the InceptionTime Score ($ITS$) and the Fr\'echet InceptionTime Distance ($FITD$) to gauge the qualitative performance of class conditional generative models on the time-series domain. We conduct extensive experiments on 80 different datasets to study the discriminative capabilities of proposed metrics alongside two existing evaluation metrics: Train on Synthetic Test on Real ($TSTR$) and Train on Real Test on Synthetic ($TRTS$). Extensive evaluation reveals that the proposed assessment method, i.e., $ITS$ and $FITD$ in combination with $TSTR$,  can accurately assess class-conditional generative model performance.}




\maketitle


\section{Introduction}\label{sec1}

In recent years, implicit generative models have gained immense popularity due to the emergence of Generative Adversarial Networks (GANs)~\cite{goodfellow2014generative}. With the astounding success of generative models in various domains such as image, video, music, and speech, it becomes imperative to quantify their performance. So far, various qualitative and quantitative assessment methods~\cite{borji2019pros, DBLP:journals/corr/abs-2103-09396} have been proposed to evaluate these models' performance and make the comparison between generative models possible. For intuitive data, there are qualitative elements like human judgment to measure the performance of a generative model. Among the quantitative methods, Inception Score ($IS$) and Fréchet Inception Distance ($FID$) have become the standard assessment methods in the image domain. Unfortunately, there is no consensual and reliable standard for evaluating generative models in the time-series domain. This deficiency impedes developing and applying deep generative models in the time-series domain and makes comparing the few existing models impossible.

Inspired by $IS$ and $FID$ from the image domain, this study introduces the InceptionTime Score ($ITS$) and Fréchet Inception Time Distance ($FITD$) to assess generative models in the time-series domain. In doing so, we investigate whether we can transfer the above-mentioned image domain standard to the time-series domain. In the literature, attempts to assess generative models have been proposed, but most notable are $TSTR$ and $TRTS$ introduced by \cite{esteban2017real}. These constitute a diametral approach compared to our $FITD$ and $ITS$ score and thus are used within our experiments to examine and control the capabilities of our newly introduced assessment metrics. 

Namely, let $P_{data}$ denote the data distribution and $P_{model}$ the distribution that our generative model learned. Ideally, we expect $P_{model}$ to be sampled from $P_{data}$ and to cover its mode space. These properties should be detectable by an assessment metric $\delta$ for it to be reliable. We designed an extensive experimental setting that includes $80$ datasets from the UCR archive\footnote{\url{https://www.cs.ucr.edu/~eamonn/time_series_data_2018/}} to investigate the quality of the sampling induced by $P_{model}$ as well as its capability to reproduce the mode space. We also involved $TSTR$ and $TRTS$ in the whole experimental pipeline and presented the efficacy so that intended researchers with an interest in conditional GANs for time-series can understand it more intuitively and gain confidence in the efficacy of the assessment metrics presented in this paper. 


\section{Related Work}
The effectiveness of generative models is normally assessed by gauging the gain in performance on the downstream task. This methodology of evaluation holds independent of the modality, i.e., image, audio, time-series, etc. Haradal et al.~\cite{haradal2018biosignal} used the improvement in a classification task to measure the quality of their generative model. Wiese et al.~\cite{wiese2019deep} described the performance of their generative model on the finance domain using statistical properties of data that are most relevant for their target domain. Another popular method is evaluating a generative model based on its performance on a surrogate task, such as supervised classification. Esteban et al.~\cite{esteban2017real} specified their generative model performance based on $TSTR$ (Train on Synthetic, Test on Real) and $TRTS$ (Train on Real, Test on Synthetic). The $TRTS$ is defined by training a classifier on real data and testing it on synthetic data. Similarly, $TSTR$ is calculated by training a classifier on synthetic data and testing it on real data. Smith et al.~\cite{smith2020conditional} employed a similar method for quantifying the performance of TSGAN.
Furthermore, the authors defined 1D $FID$ by training a simple classifier separately on each dataset and using this network for $FID$ calculation. However, the 1D $FID$ was not aligned with the visual observation from generated samples in some cases. The authors of T-CGAN~\cite{ramponi2018t} outlined the performance of their model based on the $TSTR$ only and reported AUROC instead of accuracy. While $TSTR$ and $TRTS$ can provide an indirect assessment of a generative model, they rely heavily on the choice of the classifier. Furthermore, they cannot reflect diversity in generated samples~\cite{borji2019pros}. 

\begin{figure}
    \centering
    \includegraphics[width=\textwidth]{"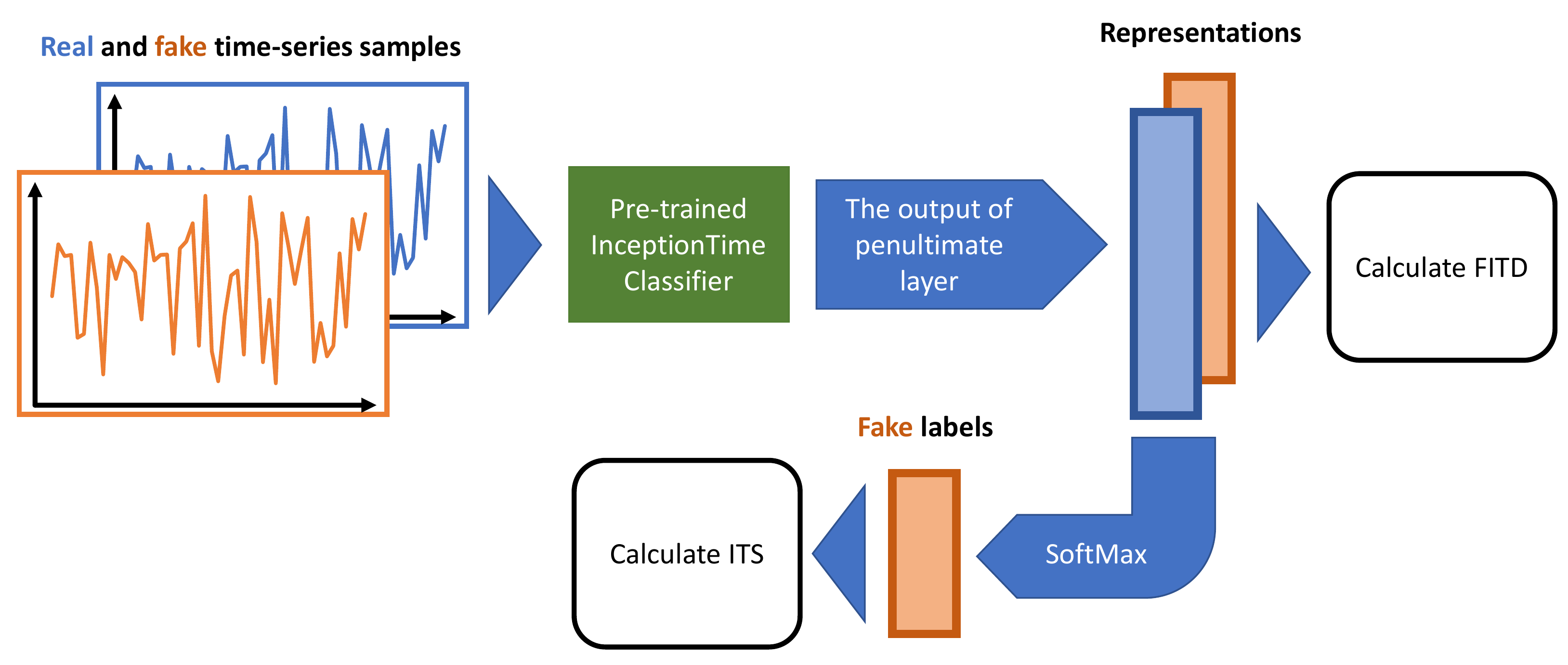"}
    \caption{The proposed evaluation pipeline for $FITD$ and $ITS$.}
    \label{fig:pipeline}
\end{figure}


\section{Quantitative Assessment for Deep Generative Models on the Time-Series Domain}
This section introduces different methods available for the assessment/evaluation of generative models with a focus on the time-series domain. All these methods employ a classifier in their pipeline, either for calculating the score or extracting features from input. To have comparable results across various studies, it is crucial to use the same testbed. For instance, on the image domain, a pre-trained inception network~\cite{DBLP:journals/corr/SzegedyVISW15} trained on ImageNet dataset~\cite{deng2009imagenet} is employed for computing assessment metrics. Therefore, in this study, we propose to adopt InceptionTime~\cite{ismail2020inceptiontime} for determining our evaluation metrics. InceptionTime is a CNN-based time-series classifier that acquired impressive accuracy on the time-series classification task. In this study, we employed a similar network structure across all datasets; however, due to the high variance between the dynamics of various time-series datasets, it is not viable to utilize a single pre-trained network across different datasets. Hence, the InceptionTime network is trained separately for each dataset. We adopt the same network structure and training pipeline as the authors of InceptionTime provided in the project git repository\footnote{\url{https://github.com/hfawaz/InceptionTime}}. An overview of our evaluation pipeline is represented graphically in Fig.~\ref{fig:pipeline}.

\subsection{InceptionTime Score ($ITS$)}
Inspired by IS for assessing generative models in the image domain, we proposed the InceptionTime Score ($ITS$) as the evaluation metric for the quality synthetic data in the time-series domain. Given $x$ as the set of synthetic time-series samples and $y$ as their corresponding labels, we expect high-quality generated data to have low entropy conditional label distribution $p(y \mid x)$. This is to be compared with the data's marginal distribution $p(y)$, which is expected to be high for diverse samples. Thus, in the ideal case, the shapes of $p(y \mid x)$ and $p(y)$ are opposite: namely narrow vs uniform. The score should reflect this property and be higher the more the conditional label and the marginal distributions differ. This is achieved by taking the exponentiation of their respective KL divergence:

\begin{equation}
\begin{split}
\text{ITS} &= \exp \left(H(y)-\mathbb{E}_{\mathbf{x}}[H(y \mid \mathbf{x})]\right) \\ &= \exp \left(\mathbb{E}_{\mathbf{x}}[\mathbb{K} \mathbb{L}(p(\mathrm{y} \mid \mathbf{x}) \| p(\mathrm{y}))]\right).
\end{split}
\label{eq:its}
\end{equation}

By definition, $ITS$ is a positively oriented metric. Its lowest value is 1.0, and its upper bound is the number of classes in the dataset. To acquire the label of synthetic time-series data, we employed a pre-trained InceptionTime network.


\subsection{Fréchet InceptionTime Distance ($FITD$)}
$ITS$ relies solely on the statistics of the generated samples and ignores real samples. Hence, it assigns a high score to a model with sharply distributed marginal and diverse training samples, regardless of whether the generated samples follow the target distribution. To address this problem on image domain, Heusel et al.~\cite{heusel2017gans} proposed Fréchet Inception Distance ($FID$). To exploit $FID$ on time-series data, we defined Fréchet InceptionTime Distance ($FITD$). We extract the feature vectors for the real and the generated samples from the penultimate layer of a pre-trained InceptionTime Classifier. We assume each of these feature vectors follows a continuous multivariate Gaussian. Subsequently, we calculate the Fréchet Distance (also known as Wasserstein-2 distance) between these two Gaussian, i.e. 

\begin{equation}
\text{FITD(r, g)}=\left\|\mu_{r}-\mu_{g}\right\|_{2}^{2}+\operatorname{Tr}\left(\Sigma_{r}+\Sigma_{g}-2\left(\Sigma_{r} \Sigma_{g}\right)^{\frac{1}{2}}\right),
\label{eq:fid}
\end{equation}

where ($\mu_{r}$, $\Sigma_{r}$) and ($\mu_{g}$, $\Sigma_{g}$) are the mean and covariance matrices of the real data and generated data, respectively. Lower $FITD$ indicates a smaller distance between data distribution and real distribution, and the minimum value is zero. $FITD$ is a robust and efficient metric; however, its assumption on multivariate Gaussian distribution in feature space is not always true.


\subsection{Assessment Based on Classification Accuracy}
We can use a classifier to explicitly benefit from labeled data to assess the class-conditional generative models. The core idea is that if a generative model can generate realistic data samples, it should perform well in the downstream tasks. In this case, a classifier can be trained on real data and tested on synthetic data in terms of classification accuracy. This paper refers to this method as $TRTS$ (Train on Real, Test on Synthetic). $TRTS$ implies that if the distribution learned by the generative model $P_{model}$ matches the data distribution $P_{data}$, then a discriminative model trained on samples from $P_{data}$ can accurately classify generated samples from $P_{model}$. $TRTS$ outputs low accuracy if generated samples fall out of $P_{data}$. However, if $P_{model} \subset P_{data}$, then $TRTS$ might assign a high accuracy, in neglection of the fact that the mode space is only partially covered by $P_{model}$.

Another classifier-based method is to train a model on synthetic data and test it on real data. We refer to this method as $TSTR$ (Train on Synthetic, Test on Real). Like $TRTS$, the $TSTR$ argues that if $P_{model} \approx P_{data}$, then a classifier trained on generated samples can score high accuracy while classifying real samples. Unlike $TRTS$, the $TSTR$ can detect the situation where $P_{model}$ partially covers $P_{data}$; however, it cannot reflect the existence of synthetic samples that do not follow $P_{data}$. In other words, $TSTR$ provide high accuracy even if $P_{data} \subset P_{model}$. This latter case is more intuitively known as an over-parametrized model. 

In this study, we employed the InceptionTime model as the classier for calculating $TRTS$ and $TSTR$.


\section{Evaluation Data - UCR Time-series Classification Archive}
The UCR archive~\cite{UCRArchive} is a collection of 128 univariate time-series datasets designed for the classification task. It thus enables us to perform our experiments on a broad spread of datasets with various properties across different domains. Furthermore, the InceptionTime model has demonstrated impressive performance in the classification task on the UCR archive. As discussed above, we need highly classifiable and diverse features to precisely calculate $FITD$ and $ITS$. Therefore, for our experimental setting, we select a subset of datasets from the UCR archive on  which the InceptionTime model acquires at least 80\% accuracy, resulting in 80 datasets. Appendix~\ref{appendix:ucr} lists the names of these datasets, their properties, and the accuracy scored by the InceptionTime model.


\section{Experiments and Results}
\label{sec:experres}

To investigate the discriminative ability of $ITS$, $FITD$, $TRTS$, and $TSTR$ in the time-series domain, we first design scenarios to replicate common problems of generative models, namely:
\begin{itemize}
    \item Decline in Quality
    \item Mode Drop, and
    \item Mode Collapse.
\end{itemize}

Subsequently, we apply our assessment methods and study how they can indicate these problems.


\subsection{Experimental Evaluation Score}
In our experiments, we train InceptionTime on the train set of these datasets and calculate our scores on the respective test set to obtain the base score ($\text{score}_{\text{base}}$) on each dataset. Since the test set is obtained from data distribution, we consider $\text{score}_{\text{base}}$ as the best score we can acquire on each dataset empirically. Also, we indicate the score of generated samples as $\text{score}_{\text{gen}}$. Finally, we define
\begin{equation}
    \text{rel(score)} = \text{score}_{\text{base}} - \text{score}_{\text{gen}}
\end{equation}
as the score of generated samples relative to the base score. We expect $\text{rel(ITS)}~\geq~0 \;$, $\text{rel(TRTS)}~\geq~0 \;$, $\text{rel(TSTR)}~\geq~0 \;$ and $\text{rel(FITD)}\leq0 \;$ in all cases. In other words, we do not expect a better score than the base score.

\begin{figure}
    \centering
    \includegraphics[width=\textwidth]{"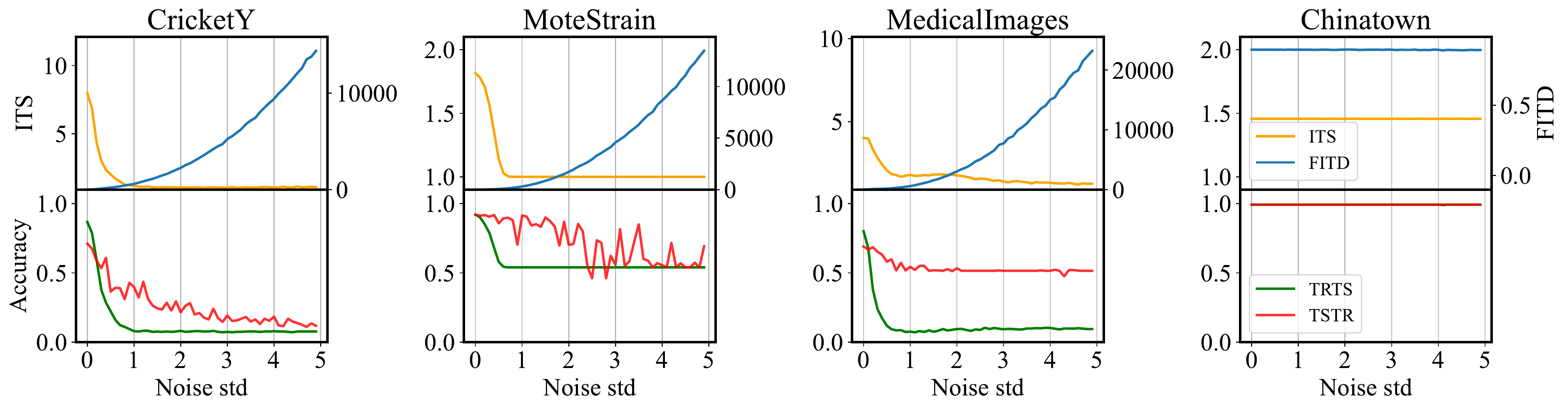"}
    \caption{Changes in the scores when data quality is declined by introducing noise into data progressively.}
    \label{fig:decline_quality_selected}
\end{figure}

\begin{figure}
    \centering
    \begin{subfigure}{0.49\textwidth}
    \centering
    \includegraphics[width=\textwidth]{"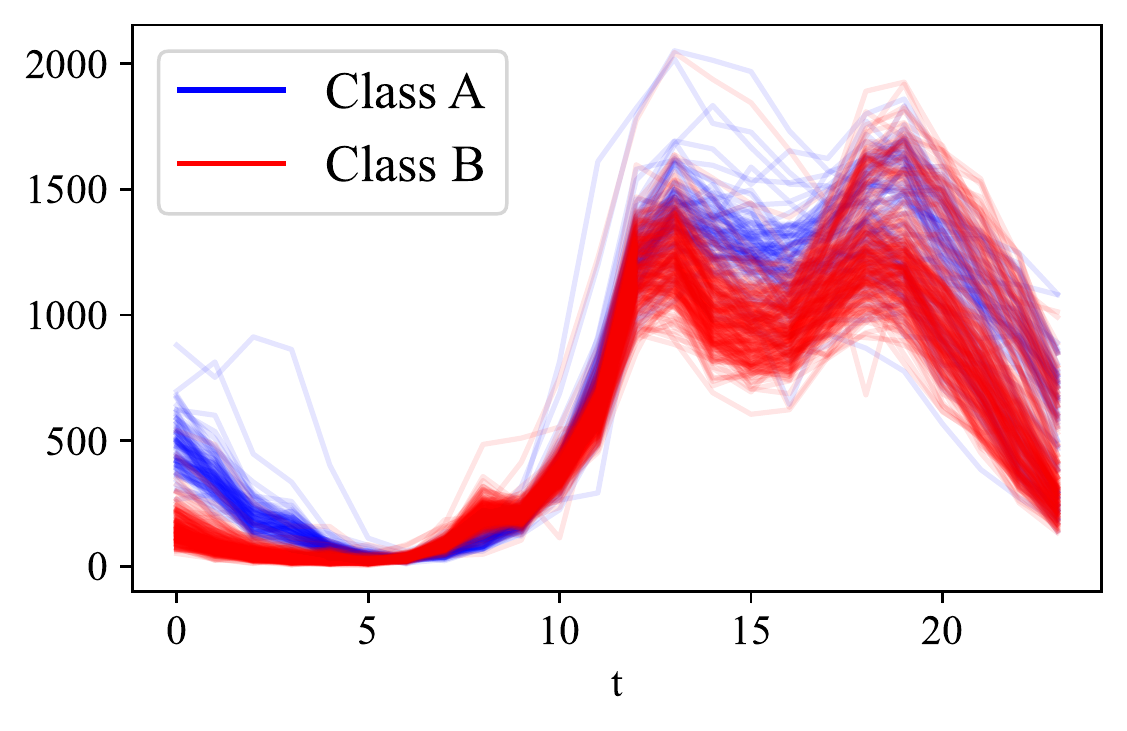"}
    \caption{The original data}
    \label{fig:china_town_a}
    \end{subfigure}
    \hfill
    \begin{subfigure}{0.49\textwidth}
    \centering
    \includegraphics[width=\textwidth]{"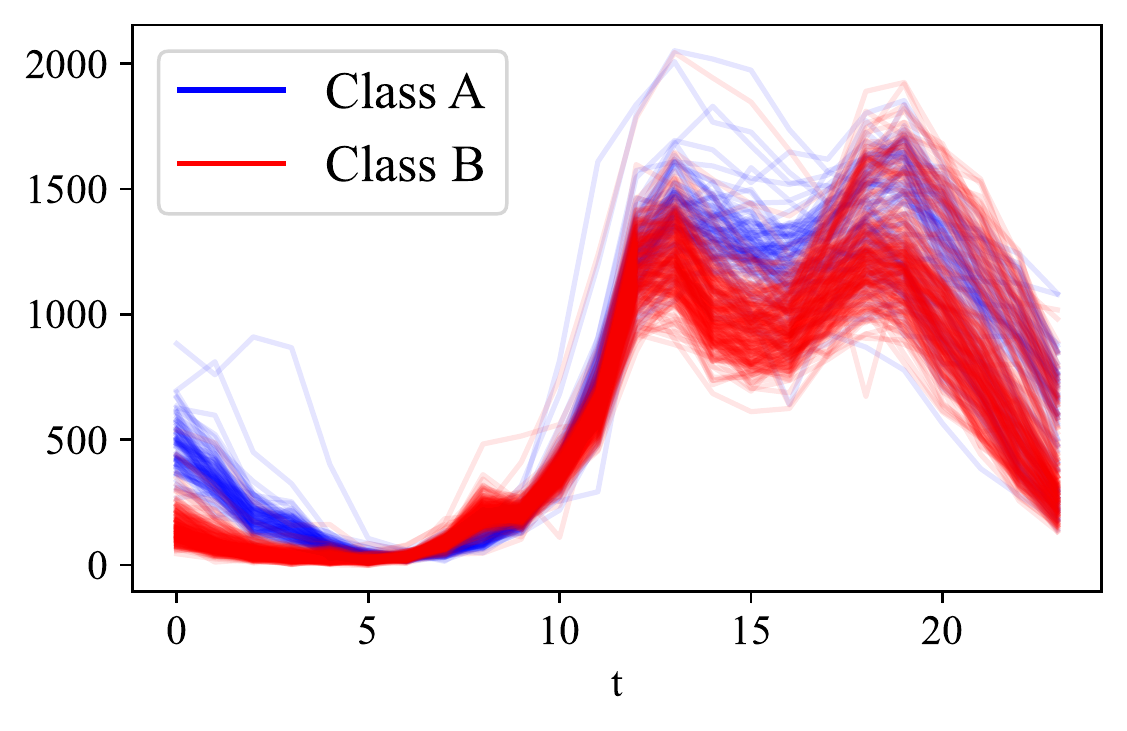"}
    \caption{Data after adding noise with $\sigma = 5$}
    \label{fig:china_town_b}
    \end{subfigure}
    \caption{The comparison between original and noisy data from the Chinatown dataset. Due to the large scale of data, the introduction of noise with $\sigma = 5$ does not change the data significantly to cause a response in our scores.}
    \label{fig:china_town}
\end{figure}


\subsection{Experiment 1 - Decline in Quality}

An assessment method should express the quality of the generated samples quantitatively. For this experiment, we added a noise signal to the samples in the test set to simulate the decrease in quality. The noise is sampled from a Gaussian distribution with $\mu = 0 \;$, and $\sigma$ is selected from an equally spaced grid of values in $[0,5]$. The standard deviation value indicates the noise strength and the amount of corruption in the original data. We expect the assessment scores to worsen with the increase in standard deviation. Figure~\ref{fig:decline_quality_selected} presents our experiment's results on four datasets (the rest of the visualization are presented in Appendix~\ref{appendix:decline}). 

\textbf{FITD}: The $FITD$ response behaves differently than others. Since $FITD$ does not have an upper bound, it increases with the increase of corruption into data. Other scores converge to their lower bound at some noise strength ($\sigma = \Delta$) and cannot indicate the increasing strength of noise on data when $\sigma > \Delta$. 

\textbf{TRTS and ITS}: The behavior of $TRTS$ and $ITS$ are very similar Both $ITS$ and $TRTS$ use the InceptionTime model trained on the train set as the backbone of their computation. Once $\sigma > \Delta$, the classifier fails to classify the samples, and its prediction is not better than a random guess. The $TRTS$ converges to random guess accuracy, which depends on the number of classes on the dataset, and $ITS$ converges to 1.0. 

\textbf{TSTR}: The $TSTR$ response has more variance than $TRTS$. The reason is that $TRTS$ is trained on a train-set of real data, which does not change during experiments, while $TSTR$ is trained on synthetic data, and as a result, we trained a new model for each value of $\sigma$.

The value of $\Delta$ depends on the scale of the data. We need more substantial noise to corrupt the data with a larger data scale. For instance, it seems that our scores cannot detect the presence of noise on data in the Chinatown data set in figure~\ref{fig:decline_quality_selected}. However, figure~\ref{fig:china_town} reveals that this data set has a great scale, ranging approximately between [0,2000]. Therefore, we need a much larger $\sigma$ to corrupt the data meaningfully.


\subsection{Experiment 2 - Mode Drop}

Mode Drop happens when the generative model ignores some modes of real data while generating artificial samples. This could be due to a lack of model capacity or inadequate optimization~\cite{arora2017generalization}. We design three experiment scenarios to evaluate the capabilities of $ITS$ and $FITD$ in recognizing mode drop in the time-series domain.

\begin{figure}
    \centering
    \includegraphics[width=\textwidth]{"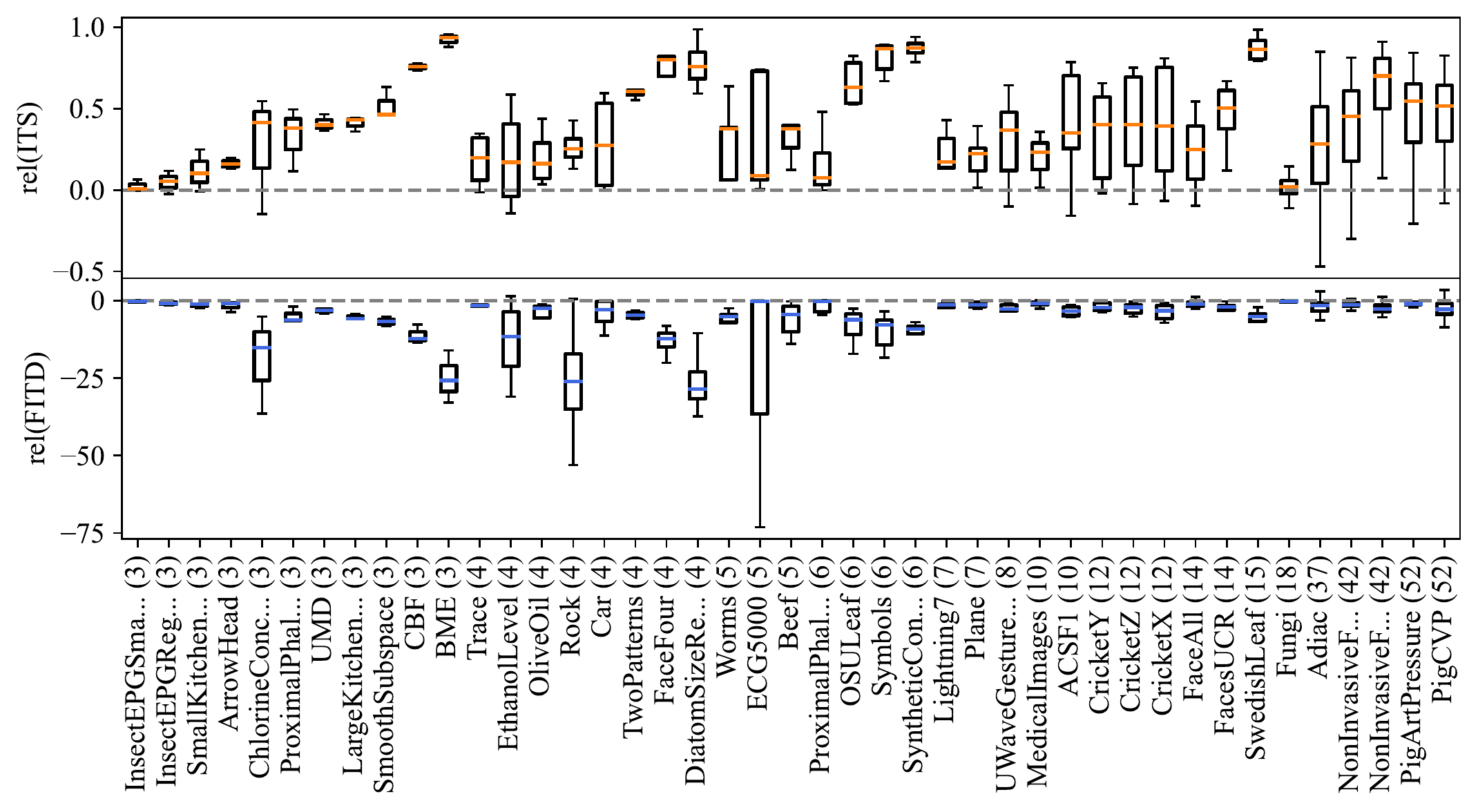"}
    \caption{Relative $ITS$ and $FITD$ score when one mode is dropped from a dataset.}
    \label{fig:mode_drop_one_is_fid}

    \centering
    \includegraphics[width=\textwidth]{"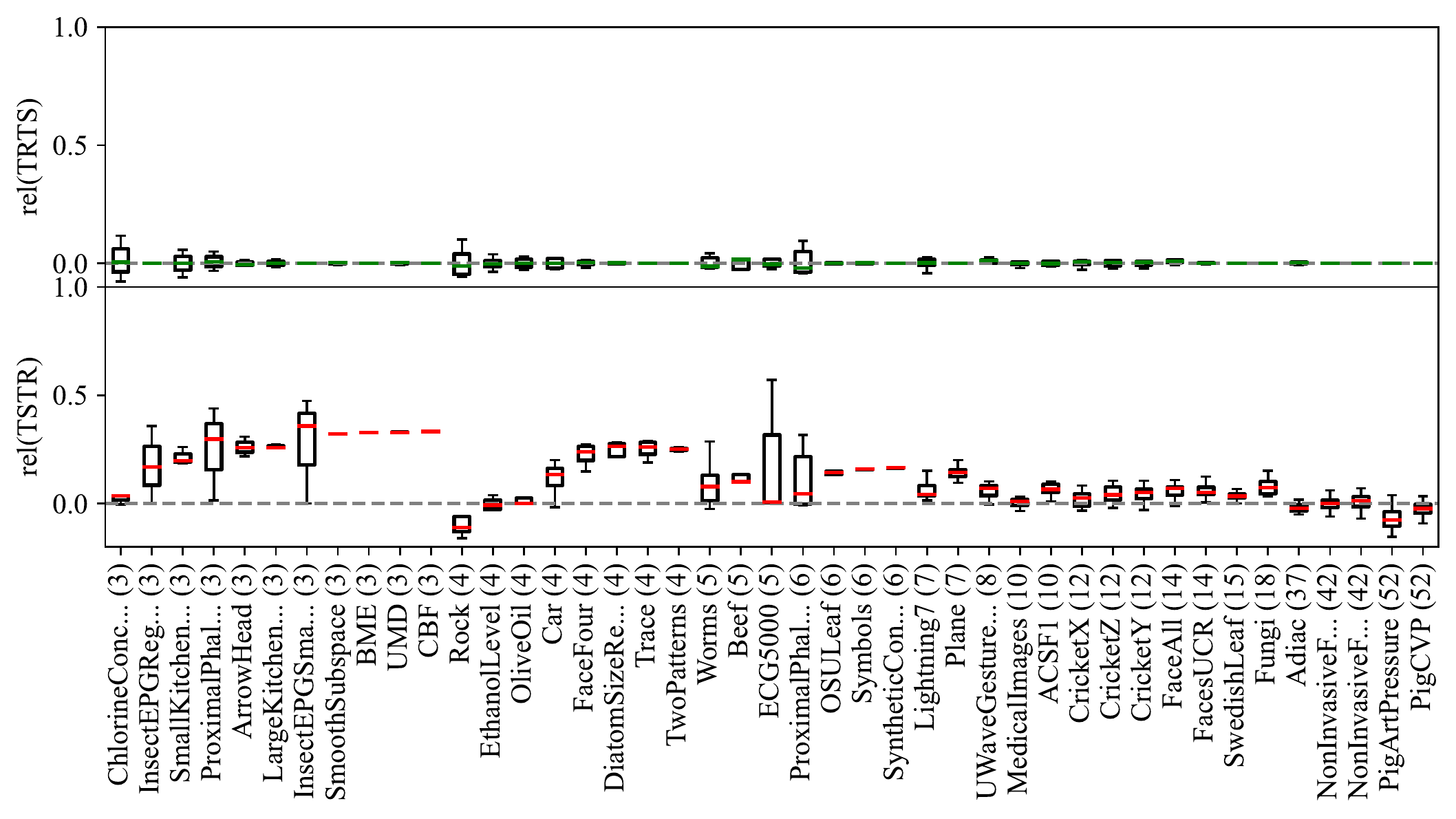"}
    \caption{Relative $TRTS$ and $TSTR$ score when one mode is dropped from a dataset.}
    \label{fig:mode_drop_one_trts_tstr}
\end{figure}

\subsubsection{Single Mode Drop}
In the first experiment, we remove all the samples belonging to one class from the test set to simulate the mode drop scenario. We calculate all scores for the mode drop caused by removing each class. Hence, for the dataset with $N$ classes, we would have $N$ values for each score. Figures~\ref{fig:mode_drop_one_is_fid} and~\ref{fig:mode_drop_one_trts_tstr} illustrate $rel(score)$ of our scores' responses on all datasets. 

\textbf{FITD}: The changes in $FITD$ depend on the degree to which the removed class affects the properties of assumed Gaussian distribution in latent space. In most datasets, the drop of a single class did not change the Gaussian distribution properties in latent space significantly. Thus, the $FITD$ reflects the single mode drop poorly. On the other hand, on a few datasets, the $FITD$ response with high variance indicates that at least one of the class samples significantly impacts the mean and covariance matrix of points in latent space. Since the feature vectors are generated with a non-linear transformation to a high dimensional space, it is impossible to interpret the $FITD$ response given the samples in the data space.

\textbf{ITS}: The $ITS$ response is mostly positive but has a great variance. When we remove a class, we change the diversity of labels. Therefore, we expect that $H(P(y \mid x))$ remains unaffected while $H(P(y))$ decreases due to the reduction in diversity. The drop of each class affects $H(P(y))$ differently, which results in a high variance between responses. If the distribution of labels is closer to a uniform distribution, the drop of each class will decrease the $H(P(y))$ similarly. In contrast, if the label distribution is heavily unbalanced, then the drop of a major class would increase $H(P(y))$. That is why we can observe the improvement in $ITS$ after mode drop in some rare cases.

\textbf{TRTS}: With the drop of a class, we have $P_{model} \subset P_{data}$. As we mentioned previously, we expect $TSTR$ to identify this situation while $TRTS$ is not capable of detecting that. Our results presented in figure~\ref{fig:mode_drop_one_trts_tstr} are aligned with these metrics' expected behavior. The $TRTS$ did not change on most datasets. 

\textbf{TSTR}: The positive $rel(TSTR)$ indicates that $TSTR$ decreases in most datasets. The impact of a single mode drop is more prominent when the dataset has fewer classes. In a few datasets, the drop of single mode has improved $TSTR$. The drop of a class has made the classification task easier for the classifier. Therefore, in a few datasets, although we have an increase in classification error due to the missing class, the classification error of other classes has been improved, which results in marginal improvement of overall accuracy.

\begin{figure}
    \centering
    \includegraphics[width=\textwidth]{"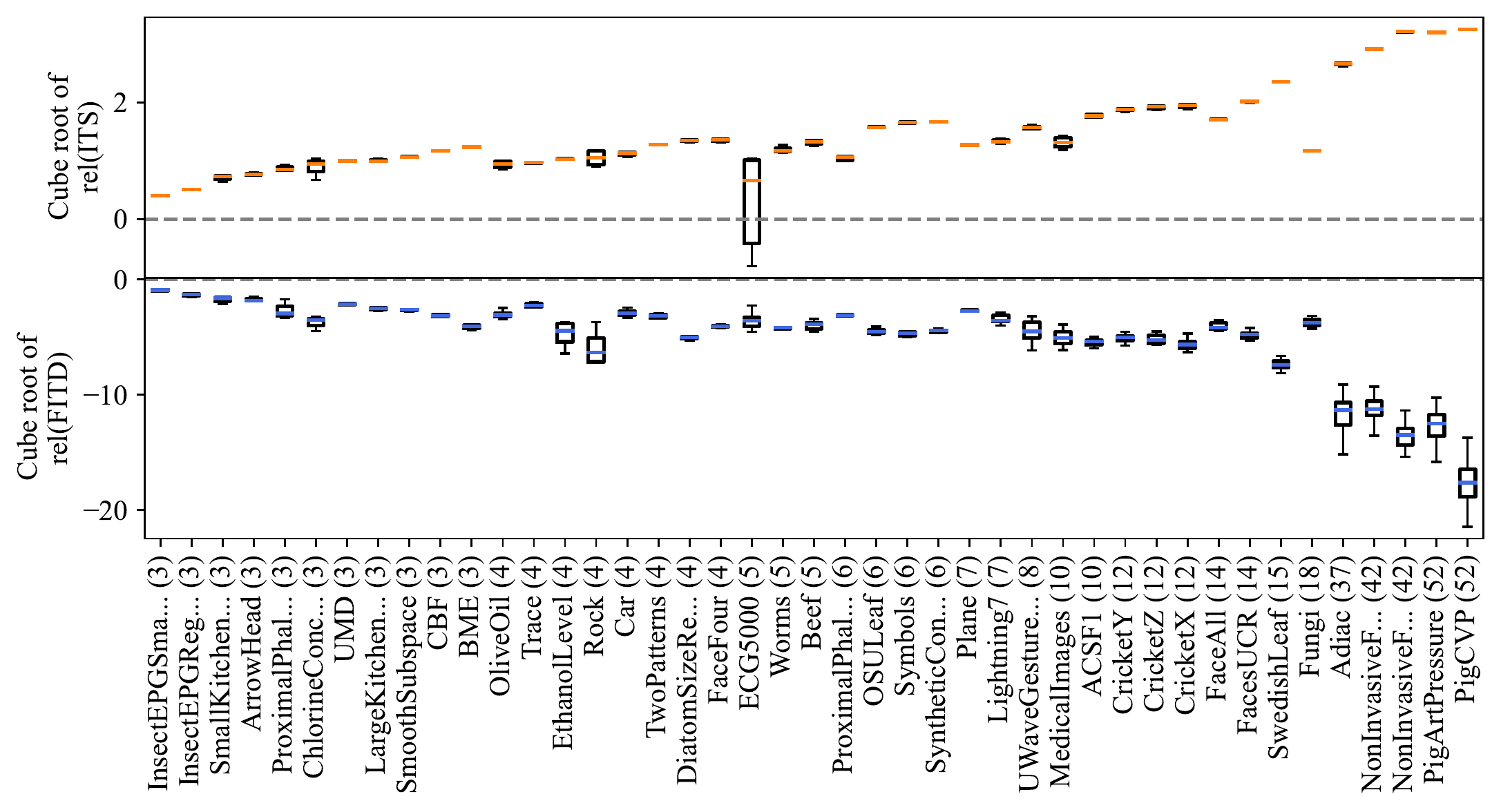"}
    \caption{Relative $ITS$ and $FITD$ score for extreme mode drop scenario. }
    \label{fig:mode_drop_extreme_is_fid}
    
    \centering
    \includegraphics[width=\textwidth]{"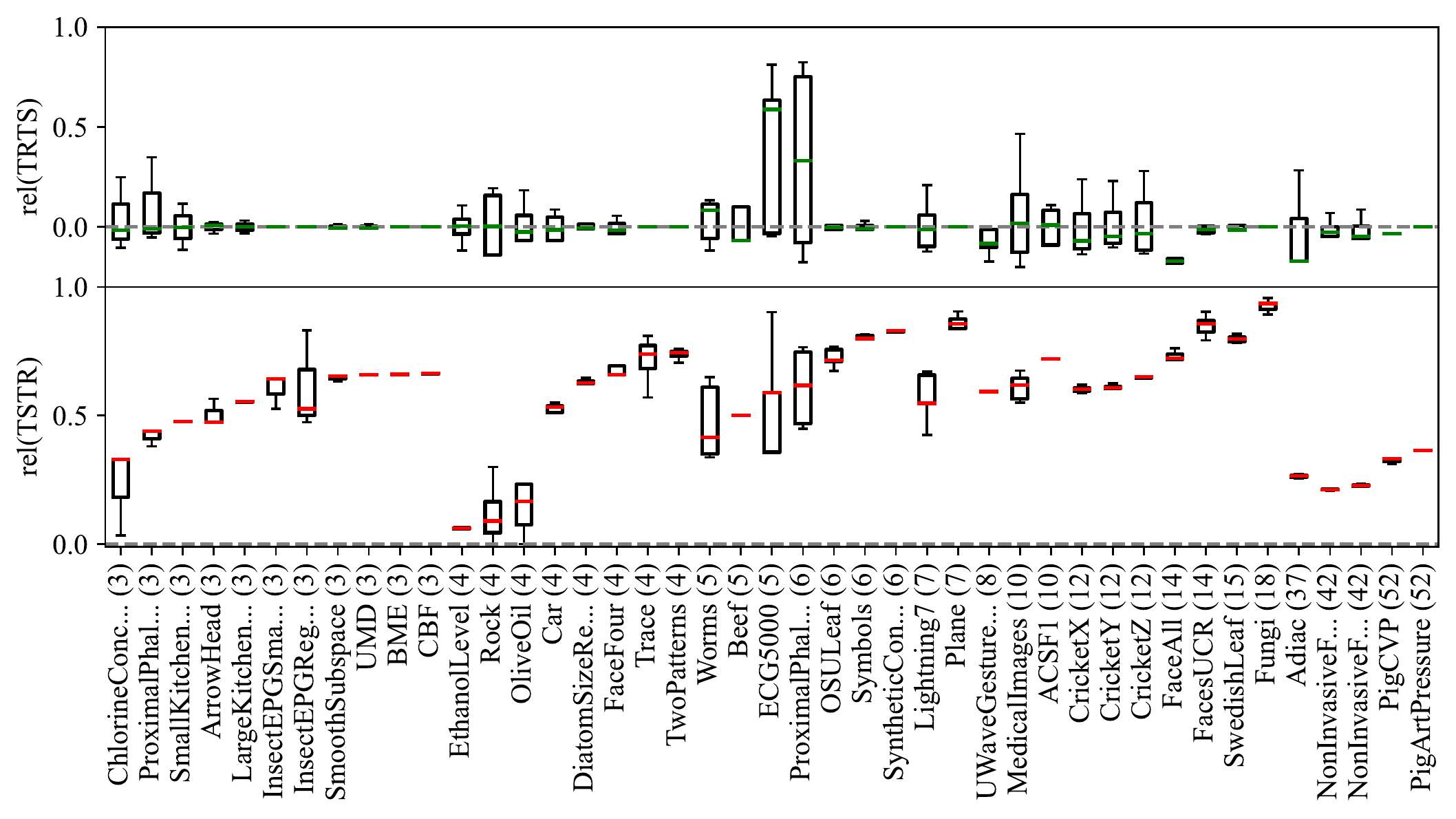"}
    \caption{Relative $TRTS$ and $TSTR$ score for extreme mode drop scenario.}
    \label{fig:mode_drop_extreme_trts_tstr}
\end{figure}

\subsubsection{Extreme Mode Drop}
In the second case, we simulate the extreme case of mode drop, where we keep only one of the classes in the test set. We follow the same approach as the previous experiment but retain only one class. Therefore, for the dataset with $N$ classes, we would have $N$ values for each score. Figures~\ref{fig:mode_drop_extreme_is_fid} and~\ref{fig:mode_drop_extreme_trts_tstr} portray the results. To make the comparison easier across all datasets, the cube root of $rel(Score)$ for $FITD$ and $ITS$ has been presented. All scores respond to the extreme mode drop scenario correctly except $TRTS$. 

\textbf{FITD}: In the case of $FITD$, the extreme mode drop drastically changes the properties of the assumed Gaussian in latent space, and we can see this shift in the $FITD$ response. Additionally, this change is more prominent with a large number of classes.

\textbf{ITS}: If we assume error-free classification, with drop of all modes except one, the $ITS = 1$ since $H(P(y)) = 0$ and $H(P(y \mid x)) = 0$ . Hence, $\; rel(ITS) = ITS_{base} - 1 = N - 1$ where $N$ is the number of classes. In practice and considering classification error, we still observe that the $ITS$ response is close to theoretical expectation.

\textbf{TRTS}: Similar to the previous experiment, $TRTS$ response cannot highlight extreme mode drop since $P_{model} \subset P_{data}$.

\textbf{TSTR}: The $TSTR$ denotes the extreme mode drop in all datasets. Furthermore, with the increase in the number of classes, we have greater divergence from $TSTR_{base}$. Please note that we have low accuracy for $TSTR_{base}$ for datasets with $N > 20$ since we trained the base model for all datasets similarly regardless of the number of classes.

\begin{figure}
    \centering
    \includegraphics[width=\textwidth]{"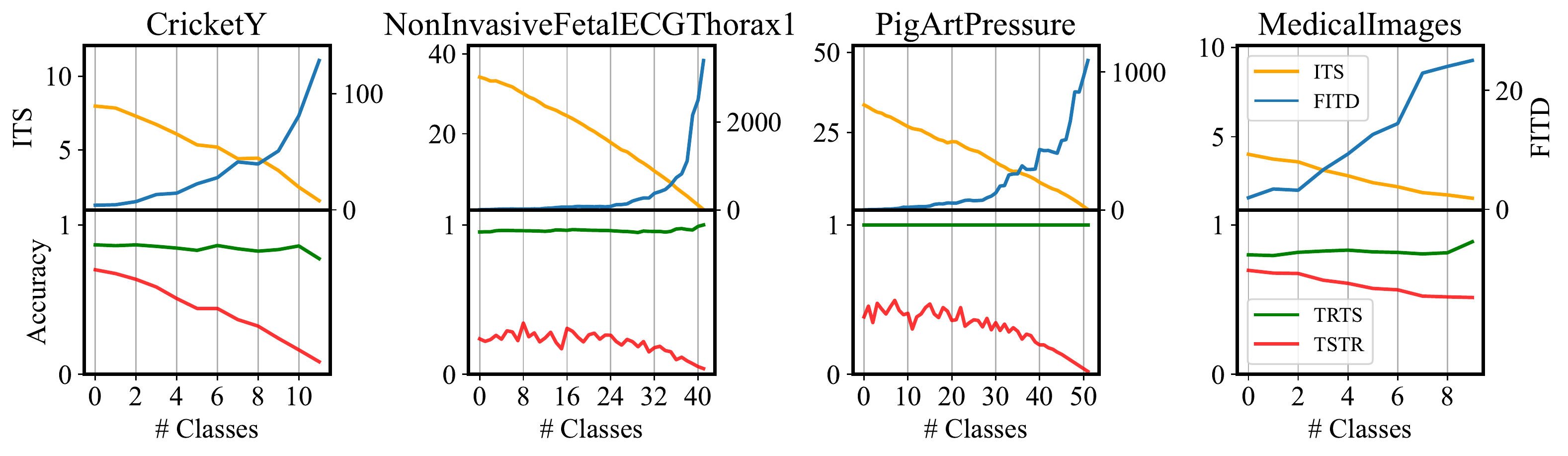"}
    \caption{Changes in the scores when modes are removed one by one.}
    \label{fig:mode_drop_increase}
\end{figure}

\subsubsection{Successive Mode Dropping}
In our final experiment, we fill the gap between the first and second experiments, drop the modes one by one, and inspect the response of our assessment method. Figure~\ref{fig:mode_drop_increase} demonstrates the scores on four datasets (the rest  of the  visualization are presented in Appendix~\ref{appendix:smd}). The results are consistent with previous experiments.

\textbf{FITD}: $FITD$ is less sensitive when a few classes are dropped. However, when the number of dropped classes crosses a certain threshold, $FITD$ increases sharply. Seemingly, the properties of assumed Gaussian distribution are quite robust against removing a few samples from the test set. However, once we remove samples belonging to most classes, the distribution begins to change dramatically with every additional class we drop from the test set.

\textbf{ITS and TSTR}: $ITS$ and $TSTR$ decrease linearly with the number of dropped classes.

\textbf{TRTS}: $TRTS$ does not change with successive drop modes.


\begin{figure}
    \centering
    \includegraphics[width=\textwidth]{"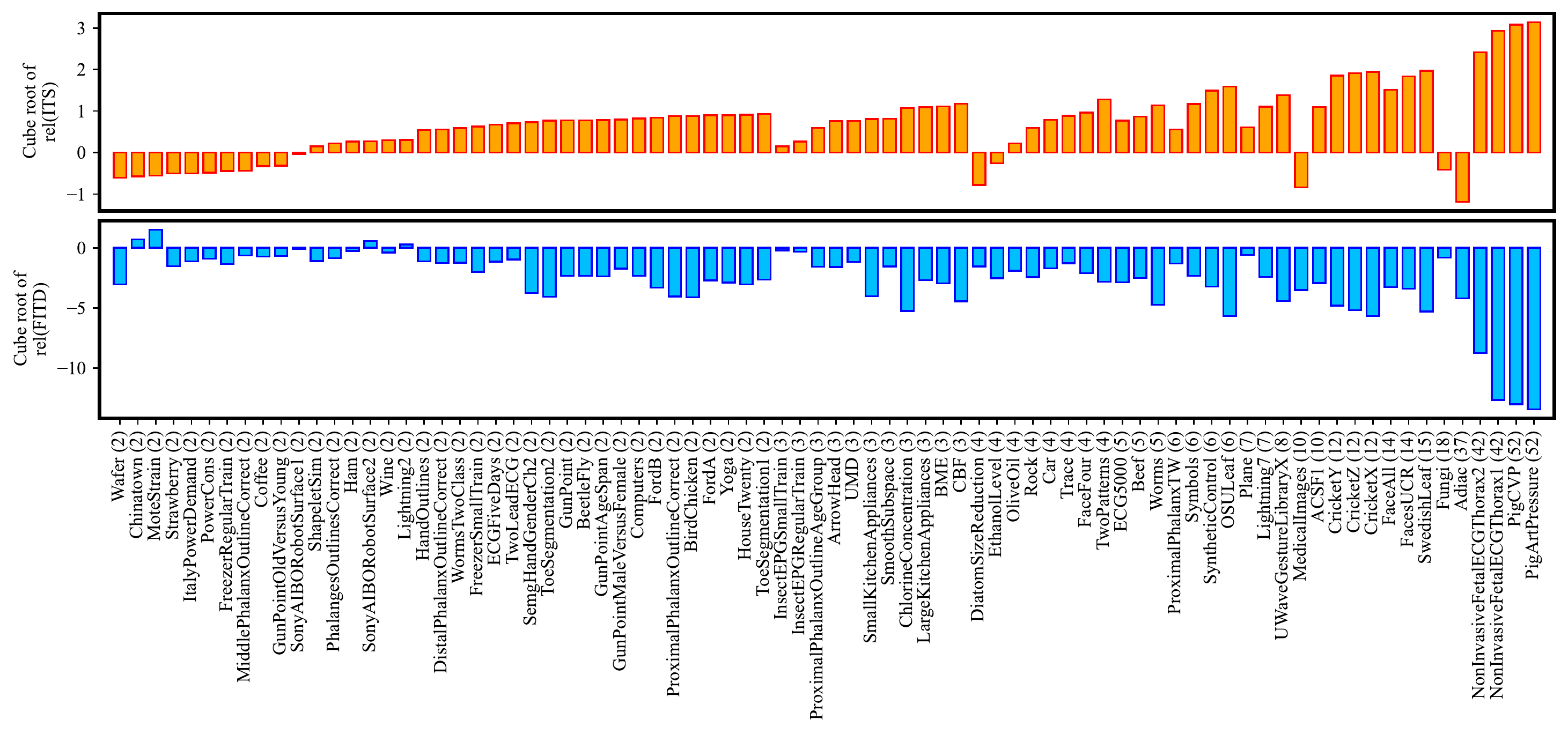"}
    \caption{Cube root of relative $ITS$ and $FITD$ score when mode collapse happens in a dataset.}
    \label{fig:mode_collapse_is_fid}

    \centering
    \includegraphics[width=\textwidth]{"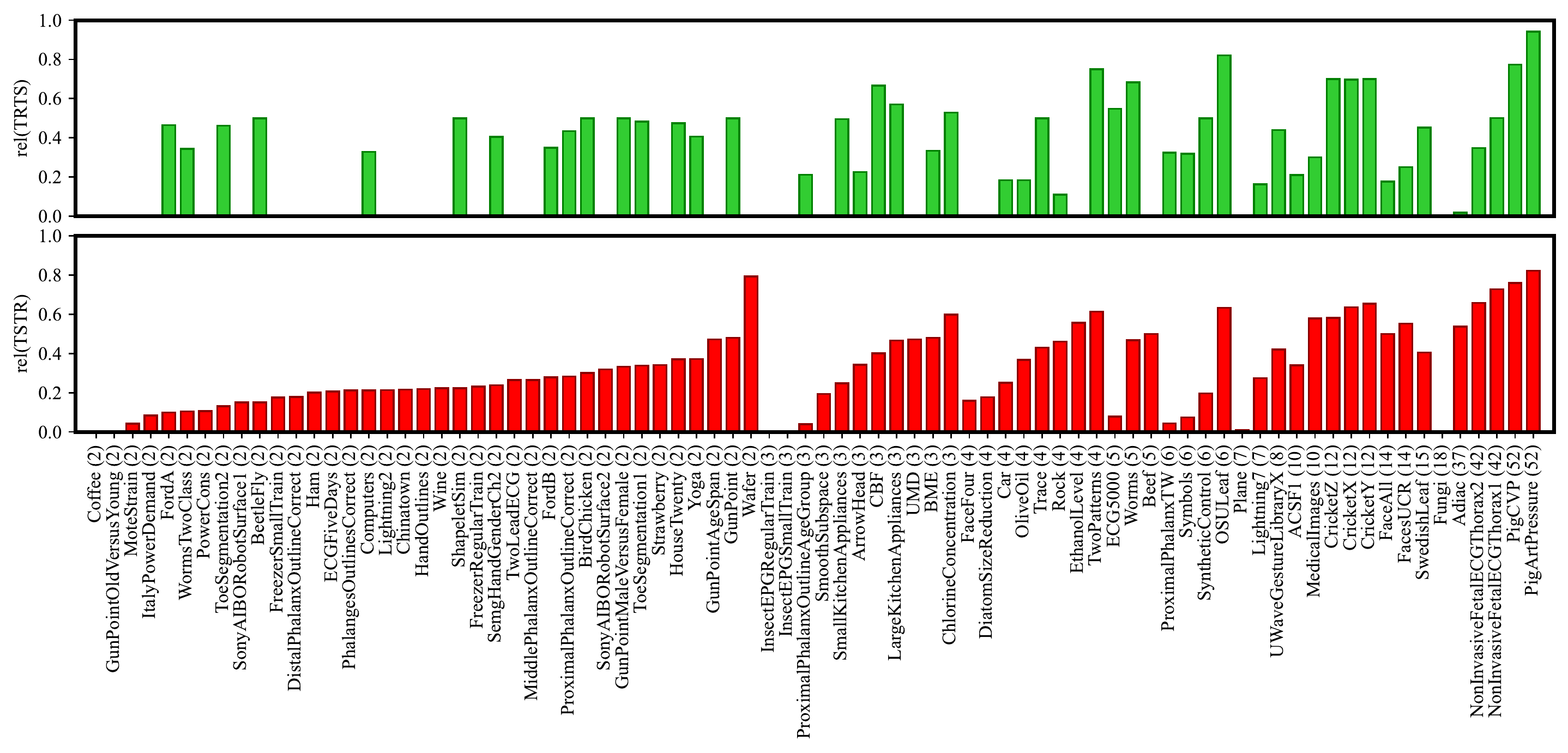"}
    \caption{Relative $TRTS$ and $TSTR$ score when mode collapse happens in a dataset.}
    \label{fig:mode_collapse_trts_tstr}
\end{figure}

\subsection{Experiment 3 - Mode Collapse}
The mode collapse problem happens when multiple modes of real data are averaged in generated data and presented as a single mode~\cite{borji2019pros}. To simulate mode collapse, we replaced samples of a class with the averaged sample. This was calculated by averaging samples in each time step as follows: Given a set of samples $\{X^{0},X^{1},...,X^{N}\}$ from a class where each sample consists of $T$ time steps ($X^{i}=\{X_{0}^{i},X_{1}^{i},...X_{T}^{i}\}$), we define the averaged sample $\overline{X}$ at time step $t \in T$ as
\begin{equation}
    \overline{X_{t}}=\frac{1}{N}\sum_{i=0}^{N}X_{t}^{i}.
\end{equation}

Figure~\ref{fig:mode_collapse_is_fid} and \ref{fig:mode_collapse_trts_tstr} summarize the performance of our scores relative to their base score in detecting this simulated mode collapse. 

\textbf{ITS}: In the presence of a perfect classifier, $ITS$ should reach its maximum since then $H(P(y \mid x)) = 0$ in \eqref{eq:its} , and we have maximum diversity among labels, hence $H(P(y)) = N \;$, where $N$ indicates the number of classes. However, the average sample might not accurately represent a class's samples. Therefore, there is a high chance of misclassification. Since our generated samples are small and are limited to a single average sample per class, any misclassification would significantly change ITS from its expected value. Therefore, we can observe in figure~\ref{fig:mode_collapse_is_fid} that the $ITS$ has been improved in some datasets.\newline
\textbf{FITD}: The $FITD$ responds correctly to mode collapse on most datasets, but its responses' strength is inconsistent across datasets. Again, interpreting the $FITD$ response depends on how samples are mapped in latent space. If the averaged samples can replicate the test set Gaussian distribution properties, we would obtain $FITD$ close to $FITD_{base}$. Otherwise, $FITD$ would diverge from its base score.\newline
\textbf{TRTS}: The $TRTS$ displays a hit-and-miss behavior. If the averaged samples can represent the original samples of the dataset, then they would classify correctly, and $TRTS$ cannot detect mode collapse. Otherwise, the misclassification of averaged samples would reflect the mode collapse problem. \newline
\textbf{TSTR}: The $TSTR$ can detect mode collapse in most datasets. When the mode collapse happens, the diversity of  generated samples decreases. Therefore, it is difficult for a classifier to learn the probability distribution of a class accurately, given only samples from the mode of the distribution. Thus, we expect a high classification error once the classifier evaluates the real data due to the limited generalization capacity of the model. The $TSTR$ behavior which is illustrated in figure~\ref{fig:mode_collapse_trts_tstr} is aligned with our expectations.


\section{Conclusion and Final Remarks}
With new advancements in the deep neural network front, the generative models are on the rise; however, their application has been hindered in the time-series domain due to the lack of a standard assessment method. In this work, we tried to alleviate this problem by introducing a framework to transform  two widely used evaluation metrics on the image domain, namely $IS$ and $FID$, to time-series. We employed the InceptionTime classifier as the backbone of our framework and introduced $ITS$ and $FITD$ for quantifying the performance of the generative model on the time-series domain. We conducted various experiments on 80 datasets to investigate the capabilities of $ITS$ and $FITD$ in detecting common problems of generative models and compare their discriminative abilities with $TRTS$ and $TSTR$, two commonly used assessment methods for class-conditional generative models. Table~\ref{tab:summ} summarizes the capabilities of these metrics in detecting three problems that generative models commonly face. Furthermore, our main findings on each metric are summarized as follows:
\begin{itemize}
  \item \textbf{ITS} can respond correctly to all the studied problems in most of the datasets; however, its behavior is most consistent in detecting the Mode Drop problem. Furthermore, $H(P(y))$ seems to be the most defining component of $ITS$ response in detecting the studied problems.
  \item \textbf{FITD} behavior heavily depends on how the samples are mapped into latent space. Since the transformation to latent space is complex and non-linear, the interpretation of the $FITD$ response is not straightforward. Additionally, since $FITD$ does not have an upper bound, it can quantify the quality of generated samples better than the other metrics.
  \item \textbf{TRTS} performance is disappointing compared to others. In the presence of other metrics, it is unnecessary to compute $TRTS$ for investigating studied problems.
  \item \textbf{TSTR} shines when the generative model has learned a subset of the real distribution. Therefore, it is the most reliable to detect Mode Drop and Mode Collapse compared to others. 
\end{itemize}

\begin{table}[]
\centering
\begin{tabular}{@{}lccc@{}}
\toprule
     & \multicolumn{1}{l}{Decline in Quality} & \multicolumn{1}{l}{Mode Drop} & \multicolumn{1}{l}{Mode Collapse} \\ \midrule
ITS  & +                                      & ++                            & +                                 \\
FITD & ++                                     & +                             & +                                 \\
TRTS & +                                      & -                             & -                                 \\
TSTR & -                                      & ++                            & ++                                \\ \bottomrule
\end{tabular}
\caption{The summary of the scores' capabilities in detecting common problems of generative models.}
\label{tab:summ}
\end{table}

This work can be extended by adopting the recent advancement of generative model assessment on image domain~\cite{DBLP:journals/corr/abs-2103-09396} to time-series domain. Another potential direction is to extend the list of studied problems or investigate other aspects of evaluation metrics such as computation time or sample efficiency.

\clearpage
\backmatter





\begin{appendices}

\section{Properties of Employed Datasets from UCR Archive}\label{appendix:ucr}
\begin{table}[h]
    \resizebox{\textwidth}{!}{
    \begin{tabular*}{1.65\textwidth}{llccc|llccc}
    \toprule
No   & Name    & Type  & Class     & Length & No   & Name    & Type  & Class     & Length\\\midrule
1	&Adiac	&Image	&37	&176	&41	&ProximalPhalanxTW	&Image	&6	&80\\
2	&ArrowHead	&Image	&3	&251	&42	&ShapeletSim	&Simulated	&2	&500\\
3	&Beef	&Spectro	&5	&470	&43	&SmallKitchenAppliances	&Device	&3	&720\\
4	&BeetleFly	&Image	&2	&512	&44	&SonyAIBORobotSurface1	&Sensor	&2	&70\\
5	&BirdChicken	&Image	&2	&512	&45	&SonyAIBORobotSurface2	&Sensor	&2	&65\\
6	&Car	&Sensor	&4	&577	&46	&Strawberry	&Spectro	&2	&235\\
7	&CBF	&Simulated	&3	&128	&47	&SwedishLeaf	&Image	&15	&128\\
8	&ChlorineConcentration	&Sensor	&3	&166	&48	&Symbols	&Image	&6	&398\\
9	&Coffee	&Spectro	&2	&286	&49	&SyntheticControl	&Simulated	&6	&60\\
10	&Computers	&Device	&2	&720	&50	&ToeSegmentation1	&Motion	&2	&277\\
11	&CricketX	&Motion	&12	&300	&51	&ToeSegmentation2	&Motion	&2	&343\\
12	&CricketY	&Motion	&12	&300	&52	&Trace	&Sensor	&4	&275\\
13	&CricketZ	&Motion	&12	&300	&53	&TwoLeadECG	&ECG	&2	&82\\
14	&DiatomSizeReduction	&Image	&4	&345	&54	&TwoPatterns	&Simulated	&4	&128\\
15	&DistalPhalanxOutlineCorrect	&Image	&2	&80	&55	&UWaveGestureLibraryX	&Motion	&8	&315\\
16	&ECG5000	&ECG	&5	&140	&56	&Wafer	&Sensor	&2	&152\\
17	&ECGFiveDays	&ECG	&2	&136	&57	&Wine	&Spectro	&2	&234\\
18	&FaceAll	&Image	&14	&131	&58	&Worms	&Motion	&5	&900\\
19	&FaceFour	&Image	&4	&350	&59	&WormsTwoClass	&Motion	&2	&900\\
20	&FacesUCR	&Image	&14	&131	&60	&Yoga	&Image	&2	&426\\
21	&FordA	&Sensor	&2	&500	&61	&ACSF1	&Device	&10	&1460\\
22	&FordB	&Sensor	&2	&500	&62	&BME	&Simulated	&3	&128\\
23	&GunPoint	&Motion	&2	&150	&63	&Chinatown	&Traffic	&2	&24\\
24	&Ham	&Spectro	&2	&431	&64	&EthanolLevel	&Spectro	&4	&1751\\
25	&HandOutlines	&Image	&2	&2709	&65	&FreezerRegularTrain	&Sensor	&2	&301\\
26	&ItalyPowerDemand	&Sensor	&2	&24	&66	&FreezerSmallTrain	&Sensor	&2	&301\\
27	&LargeKitchenAppliances	&Device	&3	&720	&67	&Fungi	&HRM	&18	&201\\
28	&Lightning2	&Sensor	&2	&637	&68	&GunPointAgeSpan	&Motion	&2	&150\\
29	&Lightning7	&Sensor	&7	&319	&69	&GunPointMaleVersusFemale	&Motion	&2	&150\\
30	&MedicalImages	&Image	&10	&99	&70	&GunPointOldVersusYoung	&Motion	&2	&150\\
31	&MiddlePhalanxOutlineCorrect	&Image	&2	&80	&71	&HouseTwenty	&Device	&2	&2000\\
32	&MoteStrain	&Sensor	&2	&84	&72	&InsectEPGRegularTrain	&EPG	&3	&601\\
33	&NonInvasiveFetalECGThorax1	&ECG	&42	&750	&73	&InsectEPGSmallTrain	&EPG	&3	&601\\
34	&NonInvasiveFetalECGThorax2	&ECG	&42	&750	&74	&PigArtPressure	&Hemodynamics	&52	&2000\\
35	&OliveOil	&Spectro	&4	&570	&75	&PigCVP	&Hemodynamics	&52	&2000\\
36	&OSULeaf	&Image	&6	&427	&76	&PowerCons	&Power	&2	&144\\
37	&PhalangesOutlinesCorrect	&Image	&2	&80	&77	&Rock	&Spectrum	&4	&2844\\
38	&Plane	&Sensor	&7	&144	&78	&SemgHandGenderCh2	&Spectrum	&2	&1500\\
39	&ProximalPhalanxOutlineAgeGroup	&Image	&3	&80	&79	&SmoothSubspace	&Simulated	&3	&15\\
40	&ProximalPhalanxOutlineCorrect	&Image	&2	&80	&80	&UMD	&Simulated	&3	&150\\\bottomrule
    \end{tabular*}}
    \vspace{1em}
    \caption{This table presents the list of selected datasets from the UCR Archive alongside their properties.}
    \label{tab:datasets}
    \end{table}
    
\begin{figure}[H]
    \centering
    \includegraphics[width=\textwidth]{"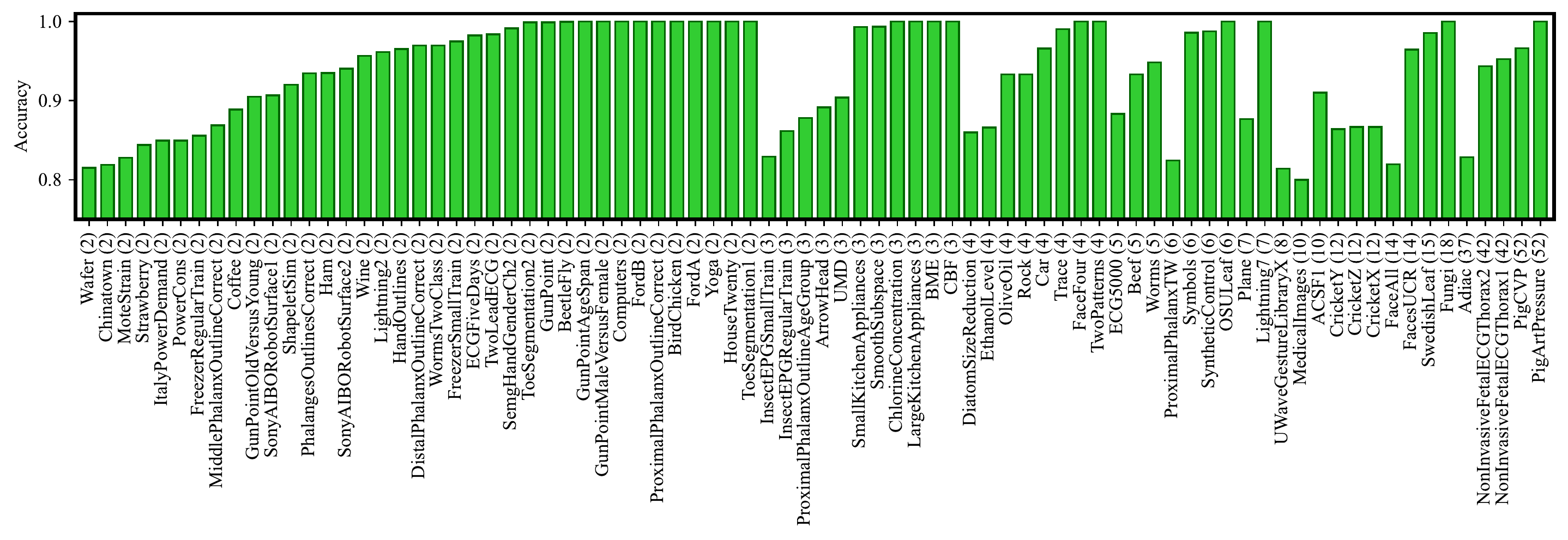"}
    \caption{The list of 80 datasets from the UCR archive alongside the accuracy of the InceptionTime classifier on these datasets. The numbers in the parentheses indicate the number of classes in the dataset.}
    \label{fig:acc}
\end{figure}
\clearpage

\section{Extra Visualization for Decline in Quality Experiment}
\label{appendix:decline}

\begin{figure}[H]
    \centering
    \includegraphics[width=0.95\textwidth]{"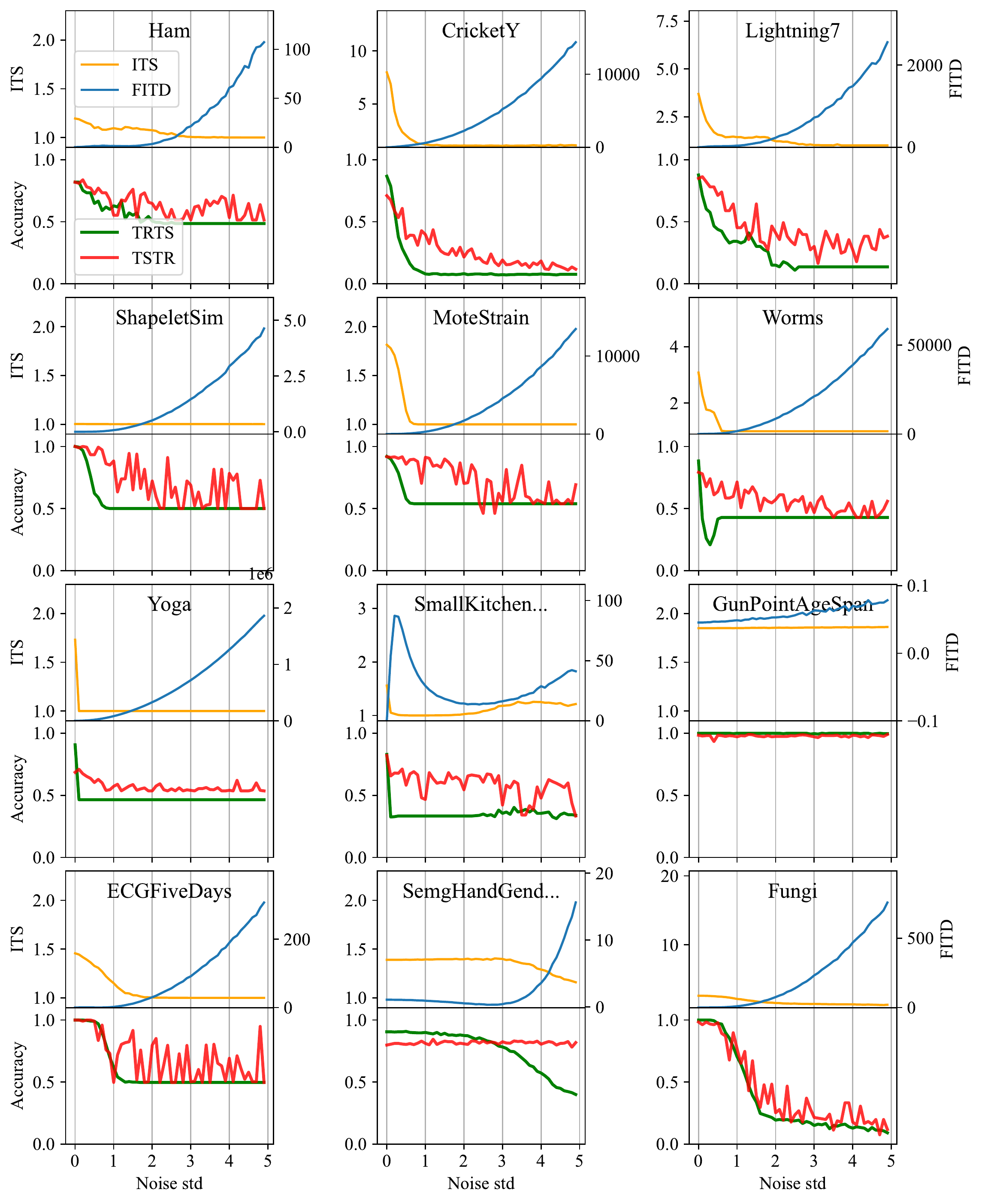"}
    \caption{Changes in studied metrics when data quality is declined by introducing noise into data progressively.}
    \label{fig:decline_quality_0}
\end{figure}

\begin{figure}[H]
    \centering
    \includegraphics[width=\textwidth]{"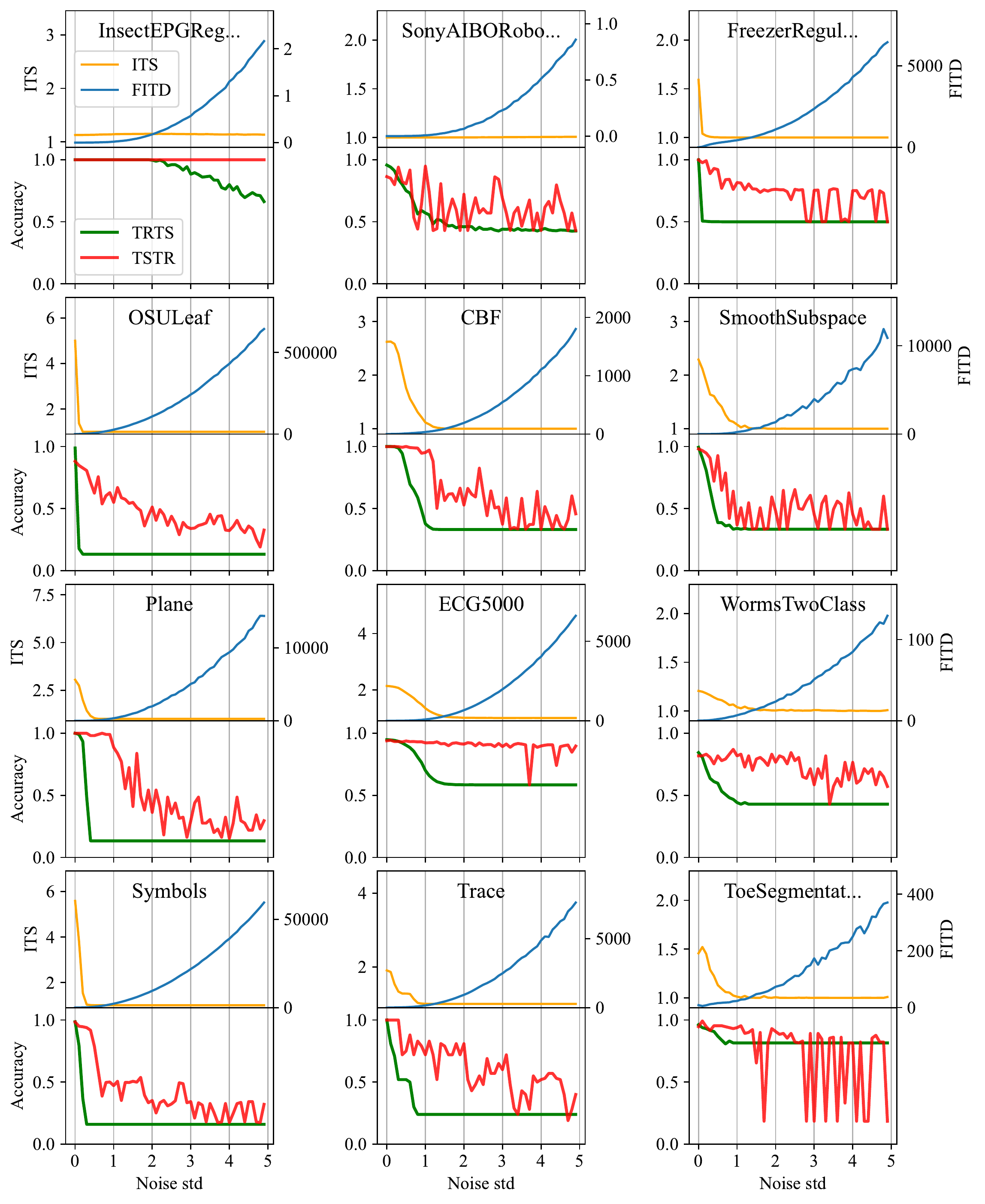"}
    \caption{Changes in studied metrics when data quality is declined by introducing noise into data progressively.}
    \label{fig:decline_quality_1}
\end{figure}

\begin{figure}[H]
    \centering
    \includegraphics[width=\textwidth]{"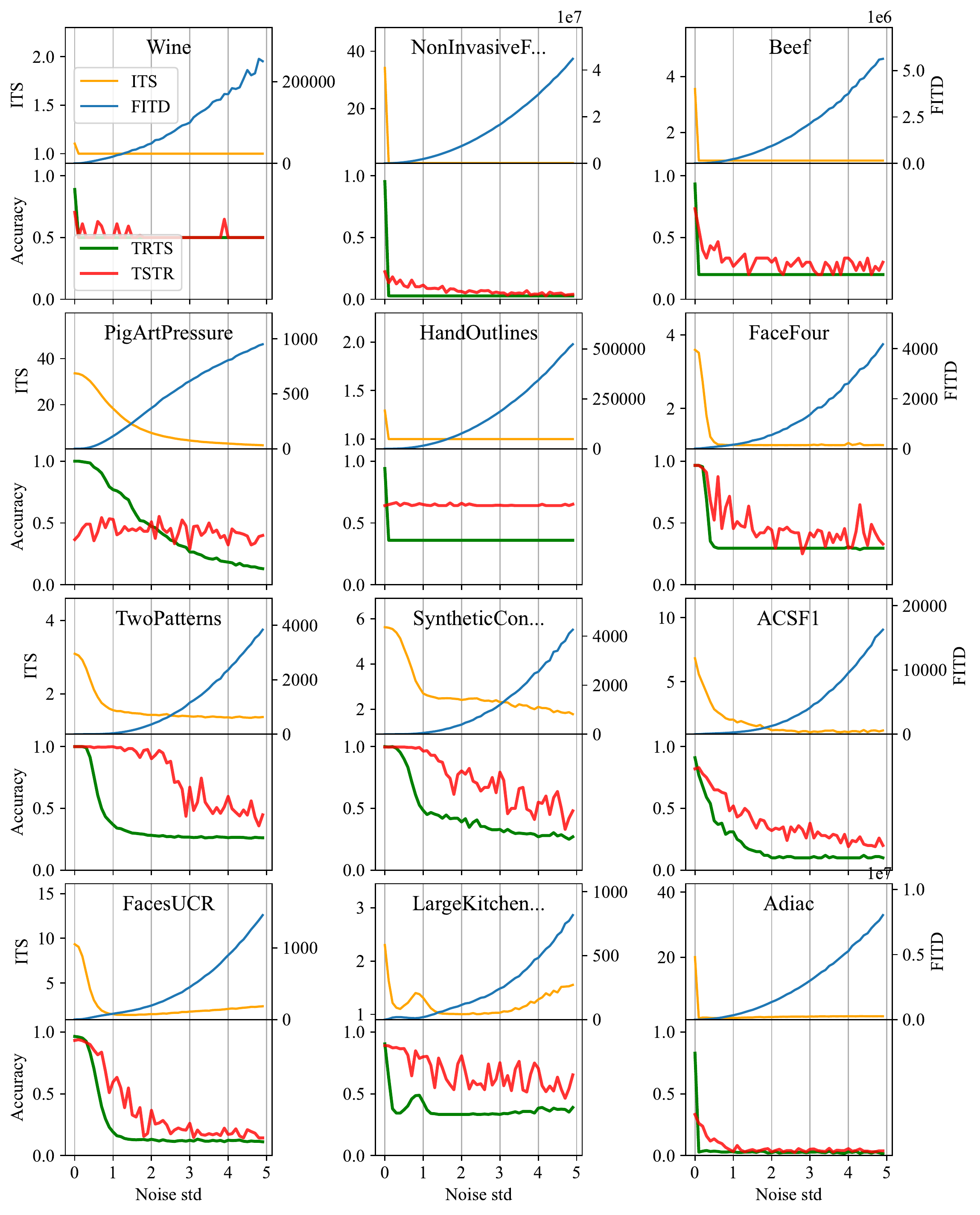"}
    \caption{Changes in studied metrics when data quality is declined by introducing noise into data progressively.}
    \label{fig:decline_quality_2}
\end{figure}

\begin{figure}[H]
    \centering
    \includegraphics[width=\textwidth]{"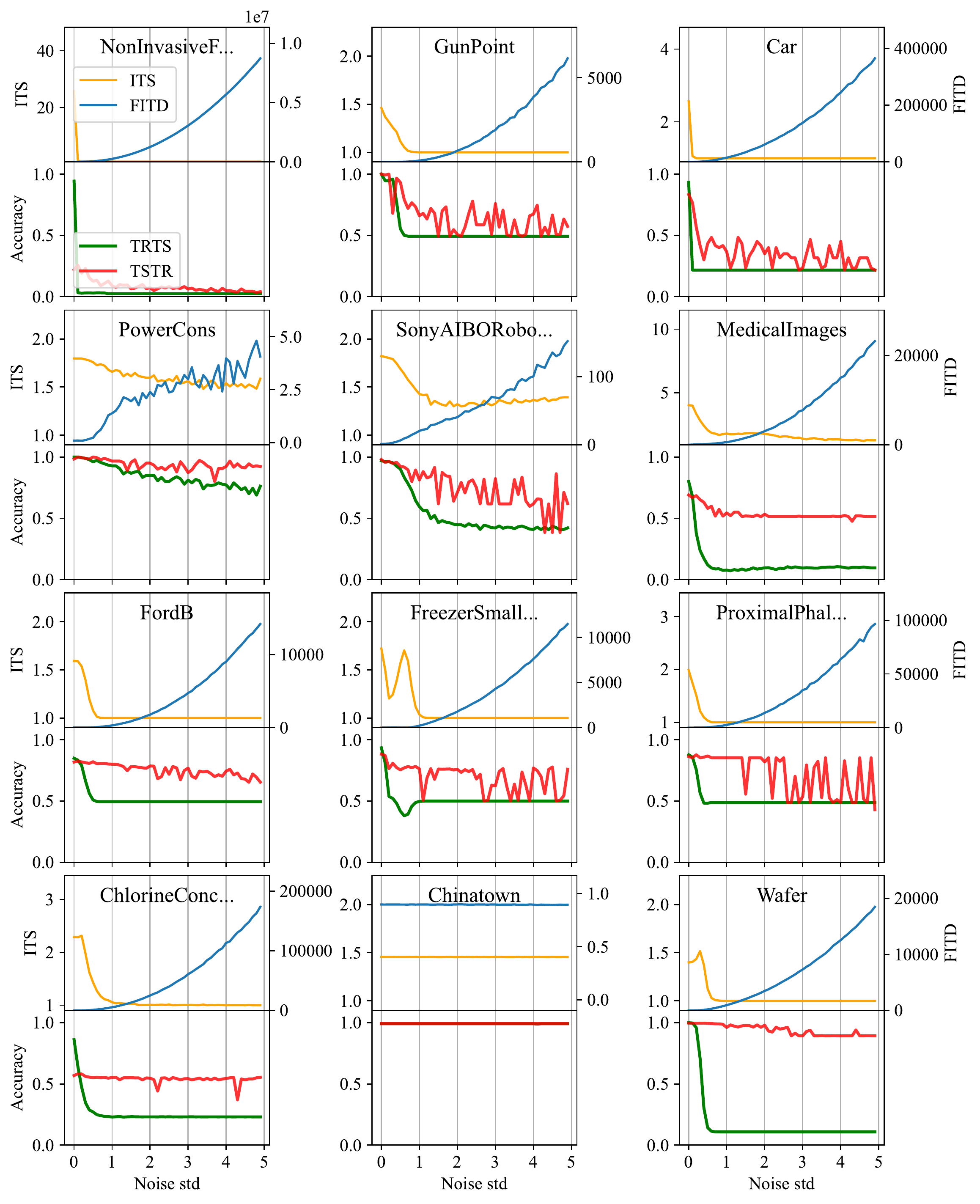"}
    \caption{Changes in studied metrics when data quality is declined by introducing noise into data progressively.}
    \label{fig:decline_quality_3}
\end{figure}

\begin{figure}[H]
    \centering
    \includegraphics[width=\textwidth]{"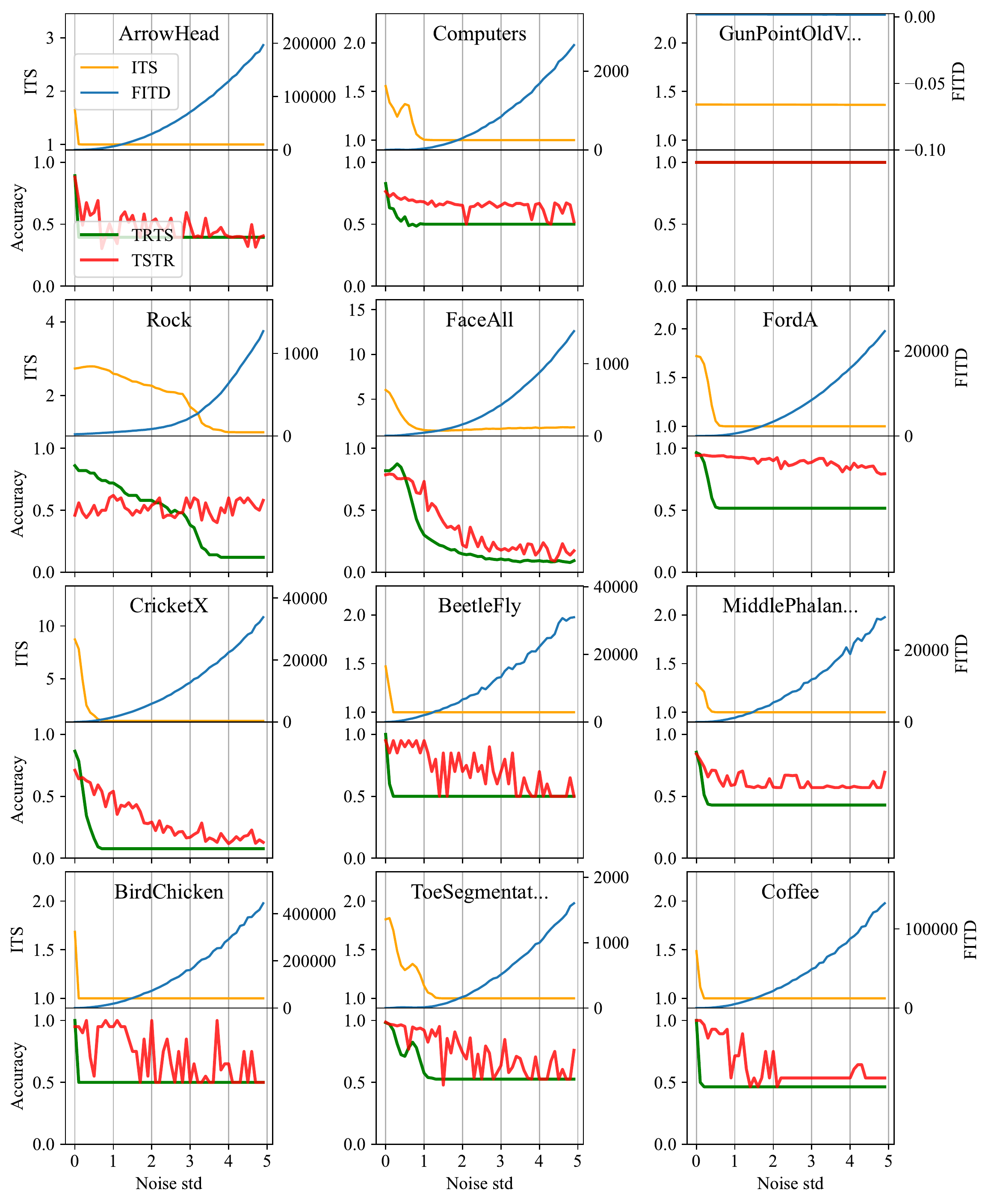"}
    \caption{Changes in studied metrics when data quality is declined by introducing noise into data progressively.}
    \label{fig:decline_quality_4}
\end{figure}

\begin{figure}[H]
    \centering
    \includegraphics[width=\textwidth]{"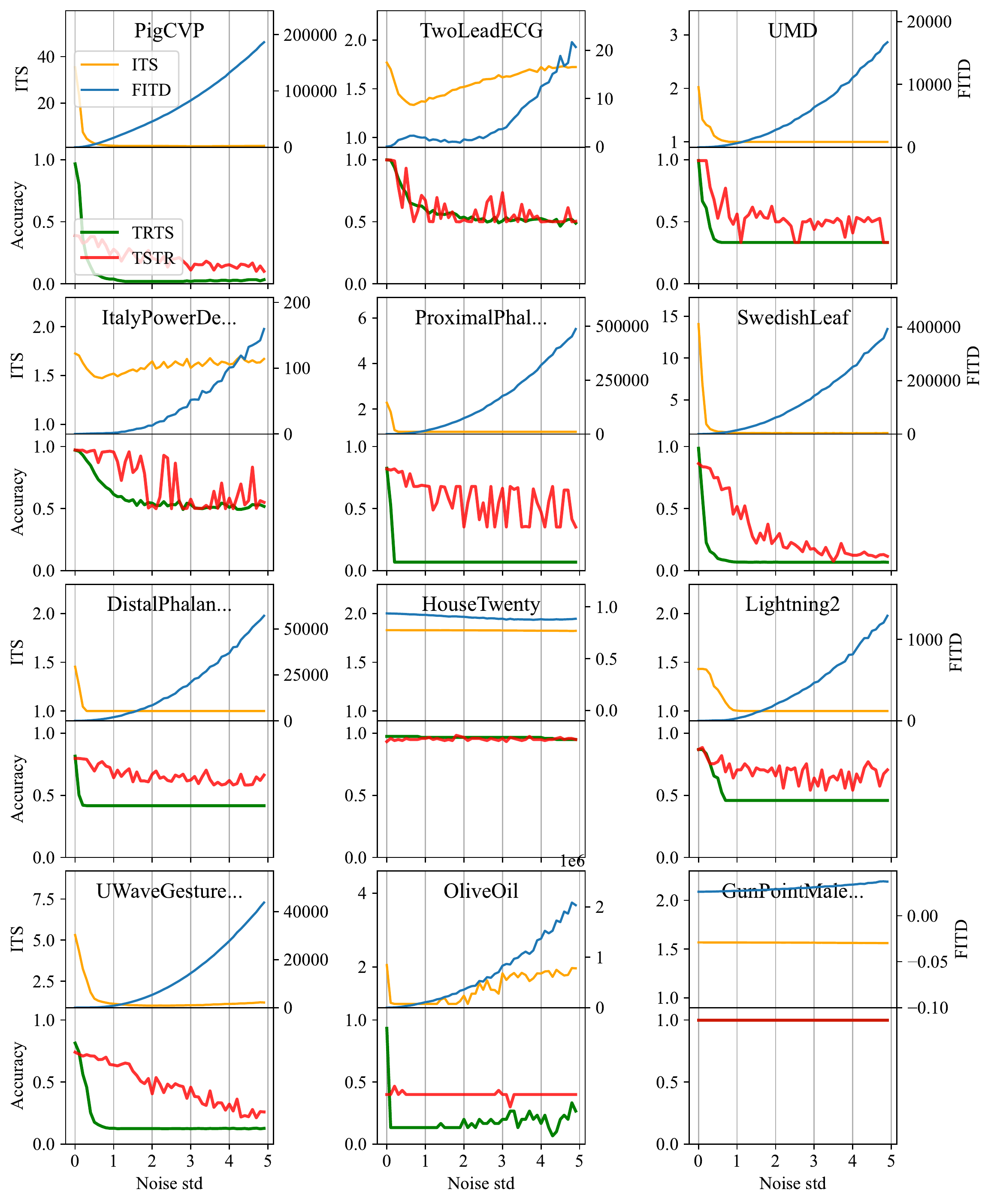"}
    \caption{Changes in studied metrics when data quality is declined by introducing noise into data progressively.}
    \label{fig:decline_quality_5}
\end{figure}

\begin{figure}[H]
    \centering
    \includegraphics[width=\textwidth]{"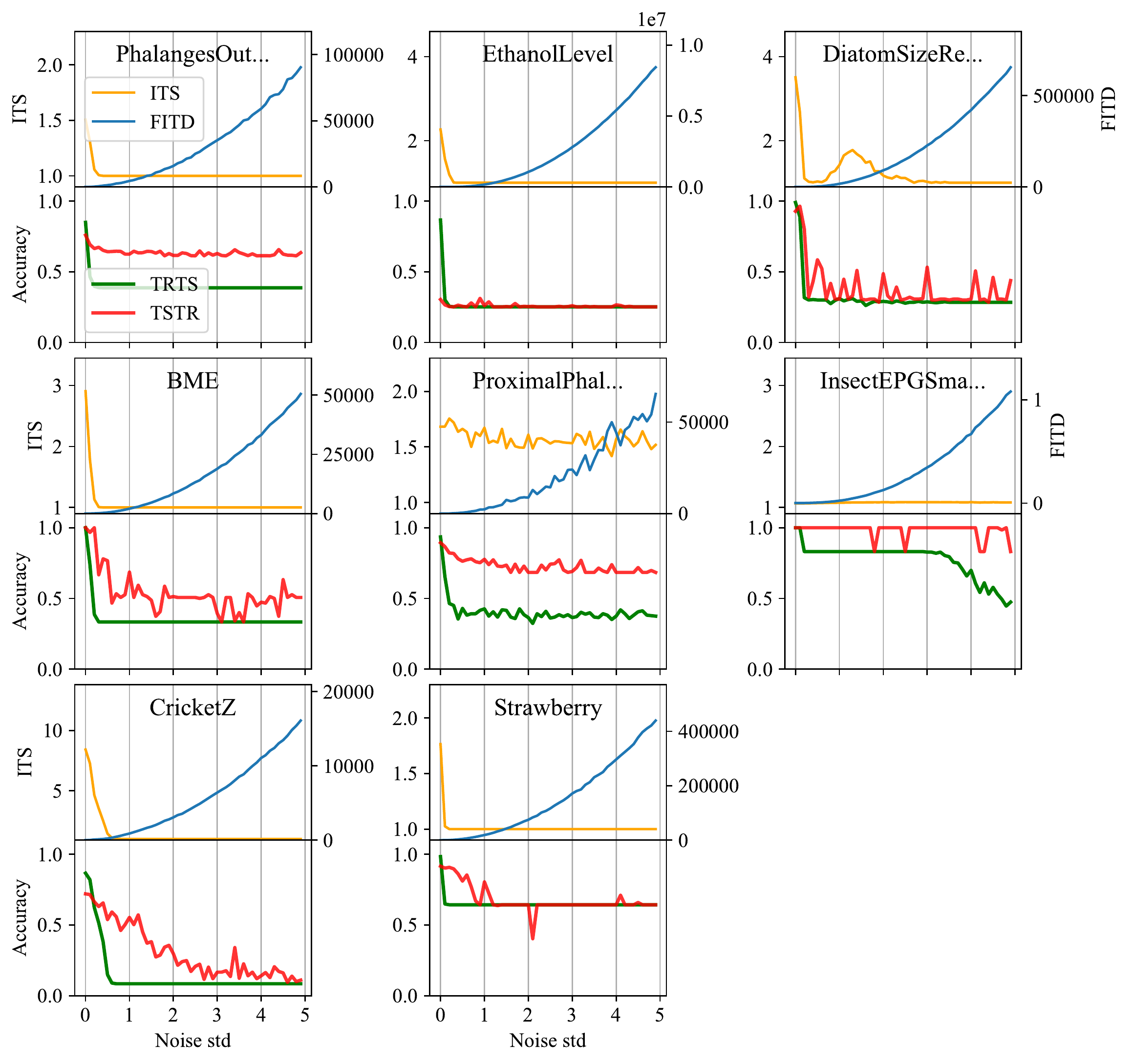"}
    \caption{Changes in studied metrics when data quality is declined by introducing noise into data progressively.}
    \label{fig:decline_quality_6}
\end{figure}

Figures~\ref{fig:decline_quality_0} to \ref{fig:decline_quality_6} provide visualization of studied metrics response for the decline in the quality experiment for all datasets in the UCR archive.

\section{Extra Visualization for Successive Mode Drop Experiment}
\label{appendix:smd}
\begin{figure}[H]
    \centering
    \includegraphics[width=0.99\textwidth]{"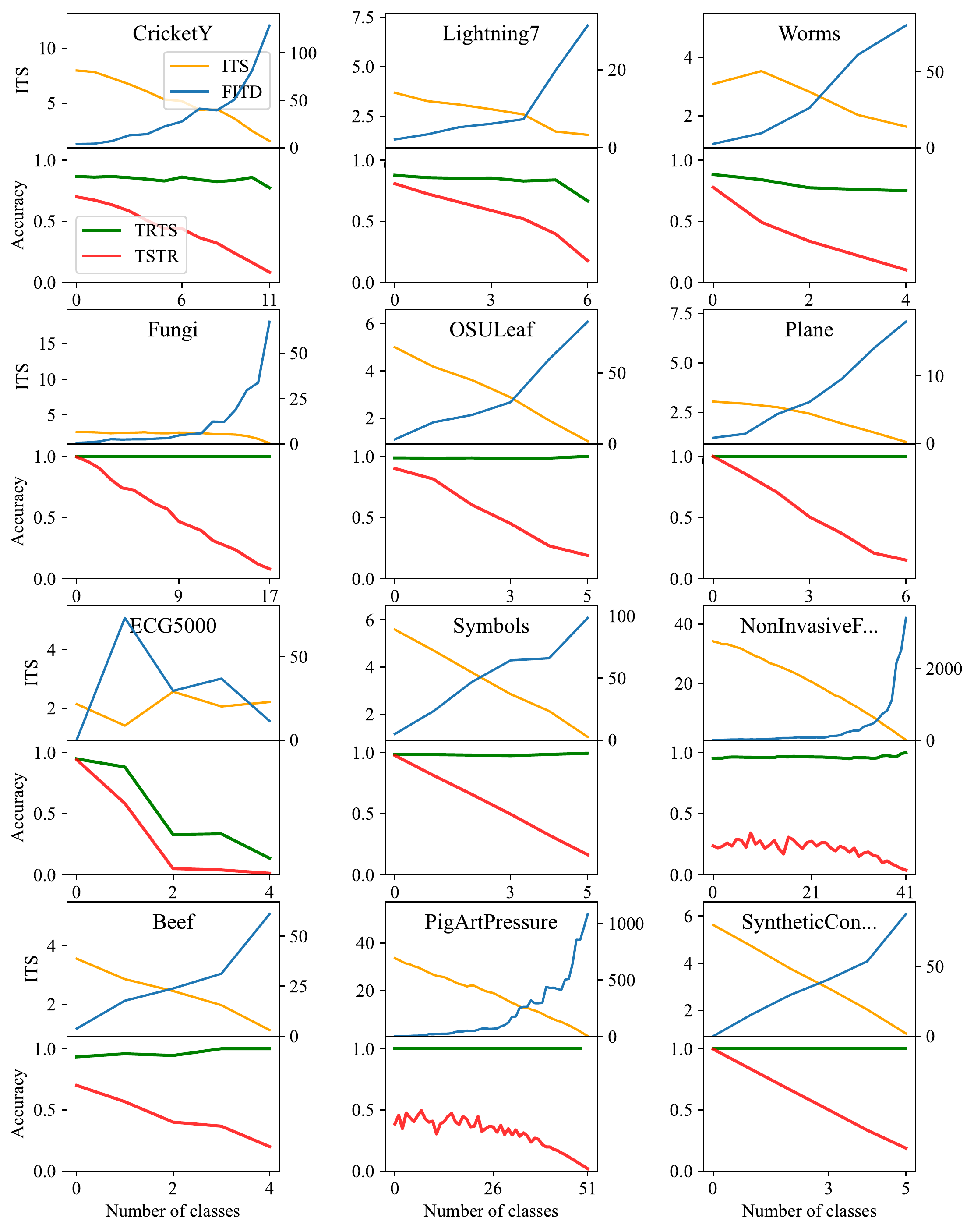"}
    \caption{Changes in studied metrics when the modes are removed one by one from a dataset.}
\label{fig:mode_drop_progressive_0}
\end{figure}

\begin{figure}[H]
    \centering
    \includegraphics[width=0.99\textwidth]{"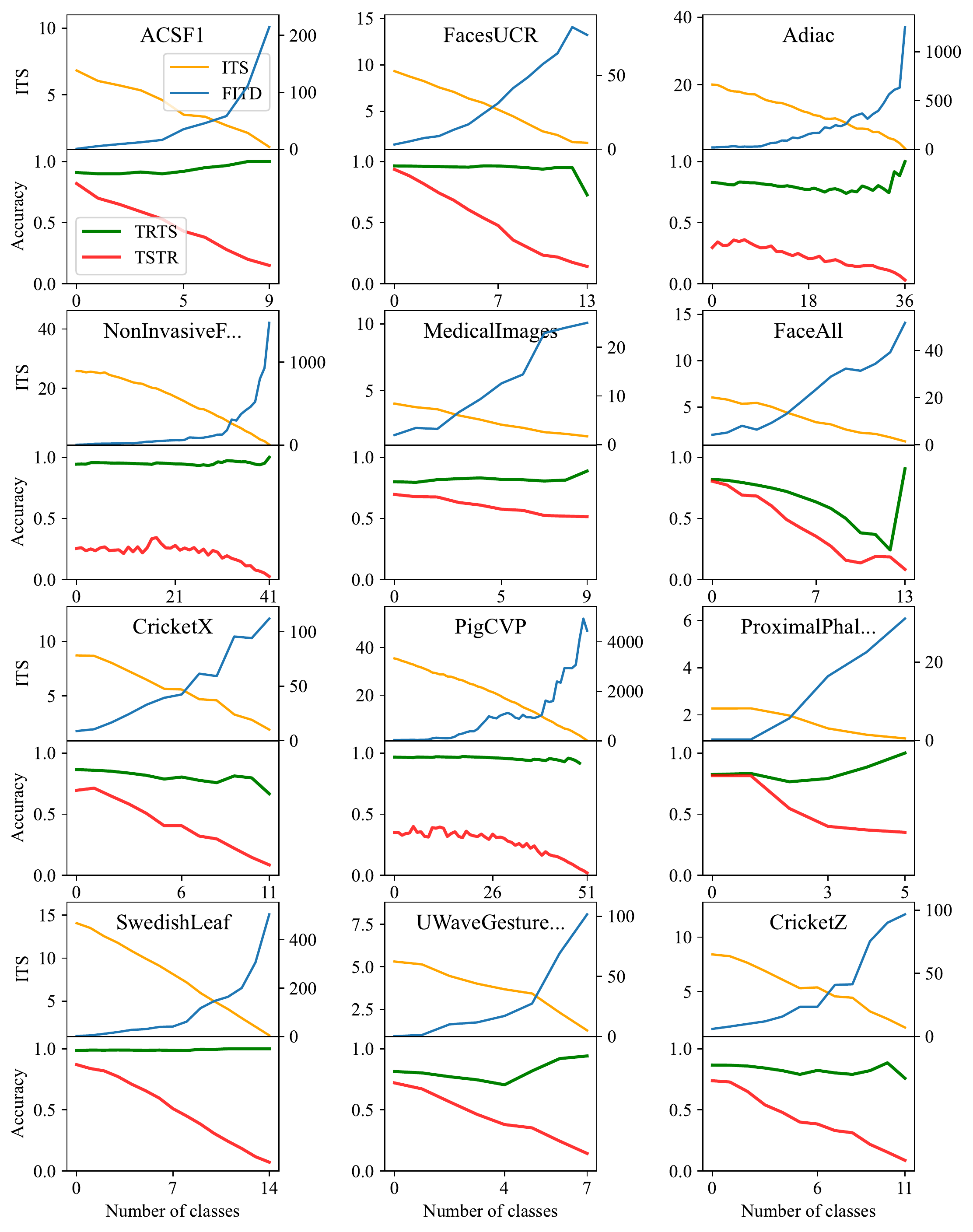"}
    \caption{Changes in studied metrics when the modes are removed one by one from a dataset.}
\label{fig:mode_drop_progressive_1}
\end{figure}

Figures~\ref{fig:mode_drop_progressive_0}, and ~\ref{fig:mode_drop_progressive_1} visualize studied metrics response when data modes are dropped progressively for datasets in the UCR archive. Only those datasets with more than five classes are presented to improve visualization.

\end{appendices}


\bibliography{sn-bibliography}


\end{document}